\documentclass[a4paper,fleqn]{cas-dc}

\usepackage[authoryear]{natbib}
\usepackage{graphicx}%
\usepackage{multirow}
\usepackage{amsmath,amssymb,amsfonts}%
\usepackage{amsthm}%
\usepackage{mathrsfs}%
\usepackage[figuresright]{rotating}%
\usepackage[title]{appendix}%
\usepackage{xcolor}%
\usepackage{textcomp}%
\usepackage{manyfoot}%
\usepackage{booktabs}%
\usepackage{algorithm}%
\usepackage{algorithmicx}%
\usepackage{algpseudocode}%
\usepackage{program}%
\usepackage{listings}%
\usepackage{comment}
\usepackage{pdflscape}

\def\tsc#1{\csdef{#1}{\textsc{\lowercase{#1}}\xspace}}
\tsc{WGM}
\tsc{QE}

\begin{document}
\let\WriteBookmarks\relax
\def\floatpagepagefraction{1}
\def\textpagefraction{.001}

\shorttitle{Incremental Affinity Propagation}    

\shortauthors{Periti et al.}  

\title [mode = title]{Incremental Affinity Propagation based on Cluster Consolidation and Stratification}



%

\author[1]{Silvana Castano}[orcid=0000-0002-3826-2407]


\cormark[4]

\ead{silvana.castano@unimi.it}

\credit{Writing - Review \& Editing}

\author[1]{Alfio Ferrara}[orcid=0000-0002-4991-4984]


\ead{alfio.ferrara@unimi.it}


\credit{Methodology}

\author[1]{Stefano Montanelli}[orcid=0000-0002-6594-6644]


\ead{stefano.montanelli@unimi.it}


\credit{Conceptualization; Writing - Review \& Editing; Supervision}

\author[1]{Francesco Periti}[orcid=0000-0001-8388-2317]

\fnmark[4]

\ead{francesco.periti@unimi.it}


\credit{Conceptualization; Methodology; Software; Validation; Investigation; Writing - Original Draft; Writing - Review \& Editing}

\cortext[1]{The authors are listed in alphabetical order}

\fntext[4]{Corresponding author}


\begin{abstract}
Modern data mining applications require to perform incremental clustering over dynamic datasets by tracing temporal changes over the resulting clusters.  
In this paper, we propose {\em A-Posteriori affinity Propagation} (APP), an incremental extension of Affinity Propagation (AP) based on \textit{cluster consolidation} and \textit{cluster stratification} to achieve faithfulness and forgetfulness. APP enforces incremental clustering where i) new arriving objects are dynamically consolidated into previous clusters without the need to re-execute clustering over the entire dataset of objects, and ii) a faithful sequence of clustering results is produced and maintained over time, while allowing to forget obsolete clusters with decremental learning functionalities. Four popular labeled datasets are used to test the performance of APP with respect to benchmark clustering performances obtained by conventional AP and Incremental Affinity Propagation based on Nearest neighbor Assignment (IAPNA) algorithms. Experimental results show that APP achieves comparable clustering performance while enforcing scalability at the same time. 
\end{abstract}


\begin{highlights}
\item Introducing the A-Posteriori Affinity Propagation (APP) algorithm for incremental clustering with cluster consolidation and stratification
\item Faithfulness and forgetfulness properties in evolutionary clustering for tracing cluster changes over time
\item Application of incremental clustering to Semantic Shift Detection and diachronic document corpus analysis
\item Comparison with benchmark algorithms, namely Affinity Propagation (AP) and Affinity Propagation based on Nearest Neighbor Assignment (IAPNA), and discussion on scalability benefits 
\end{highlights}


\maketitle

\section{Introduction}\label{sec:introduction}
The capability to perform incremental clustering over dynamic datasets is getting more and more importance in current data mining applications, like for example social network analysis~\citep{arpaci2021evolutionary}, 
climate change studies~\citep{gunnemann2012tracing}, and medical images segmentation~\citep{si2022breast}. However, conventional clustering algorithms are mostly conceived to deal with static datasets, where all the objects are available as a whole and clustering is performed offline over the entire set of data~\citep{leilei2014incremental}. Extensions based on incremental solutions are proposed to deal with dynamic datasets, where objects continuously arrive, and clustering is performed by processing new data as they appear. Instead of recomputing the clustering result from scratch every time new objects are received, \textit{incremental clustering} algorithms aim to efficiently update the clustering result by processing and assimilating the new objects into the existing clusters.

For instance, extensions for incremental clustering have been proposed for k-means and Affinity Propagation (AP) algorithms, where the focus is to find the best solution for assimilating new incoming objects into the current clustering result, rather than recomputing a new clustering result from scratch~\citep{frey2007clustering,leilei2014incremental,sunmood2018evolution,arzeno2017evolutionary}. A weighted AP extension has been proposed to deal with data streams, based on a compact description of the data flow and on the use of a reservoir where to place stream objects showing low affinity with existing clusters~\citep{zhang2008frugal}. To work with dynamic datasets, scalability issues become also relevant in designing incremental clustering algorithms, in that they have to cope with high data volumes, sequential access, and dynamically evolving nature of the data to be classified.

To support temporal evolution analysis and to trace cluster changes over time, evolutionary incremental clustering algorithms have been proposed which produce a sequence of clustering results, one for each time period~\citep{beringer2006online,hruschka2009survey,nordahl2021evolvecluster}. Two main issues become relevant in evolutionary clustering. A first issue regards the {\em faithfulness} property, that is, the clustering at any point in time should remain faithful to the current data as much as possible, thus avoiding resulting clusters to dramatically change from one time-step to the next~\citep{chakrabarti2006evolutionary}. This property facilitates the exploitation of clustering results over time, namely the capability to trace the {\em cluster history}, since users get progressively familiar with results and can compare clustering of different time periods in a more effective way. A second issue regards the so-called {\em stability-plasticity dilemma}, that is, the phenomenon by which \textit{some patterns may be lost to learn new knowledge, and learning new patterns may overwrite previously acquired knowledge}~\citep{yang2013incremental}. Thus, faithfulness is enforced in evolutionary clustering to learn new information without forgetting what has been previously learned. As an additional property, \textit{forgetfulness} is required to discard information become obsolete, thus reducing memory usage and enforcing scalability.

In this article, we propose the {\em A-Posteriori affinity Propagation} (APP) algorithm.  APP has been initially conceived for application to Semantic Shift Detection with successful experimental results~\citep{periti2022done}. In this paper, we formalise the APP definition as an incremental extension of AP  based on \textit{cluster consolidation} and \textit{cluster stratification} to achieve faithfulness and forgetfulness. APP enforces incremental clustering in that i) new arriving objects at time $t$ are dynamically assimilated into previous cluster results without re-calculating clusters at time $t-1$ and ii) a faithful sequence of clustering results is produced and maintained over time (i.e., cluster history), while allowing to forget obsolete clusters. Cluster consolidation means that APP keeps memory of clustering results at time $t-1$ by collapsing each cluster into a summary representation, namely the {\em centroid}, which is considered as an additional object to cluster at time $t$. Cluster stratification means that the new clusters at time $t$ are obtained from clusters at time $t-1$ i) by creating a new cluster including new objects arriving at time $t$ (\textit{stratification-by-creation}), ii) by inserting new objects arriving at time $t$ into an existing $t-1$ cluster (\textit{stratification-by-enrichment}), iii) by merging two or more $t-1$ clusters into a new one at time $t$  (\textit{stratification-by-merge}).

APP can be used for discovering concepts in incremental scenarios under the assumption of  ``group evolution'', in contrast to the ``individual evolution''. A new incoming object dissimilar from the past observations tends to be considered by APP as an outlier of a previously generated cluster rather than a unique exemplar of a new cluster. This means that a new cluster can be detected only if there is a relevant number of incoming exemplars associated with it.
Finally, to enforce forgetfulness, a decremental learning functionality is defined in APP to allow the selective pruning of aged, obsolete clusters, similarly to the {\em forgetful property of human mind}~\citep{yang2013incremental}. 

For evaluation, we consider popular labeled datasets and we compare APP against benchmark incremental clustering algorithms (i.e., AP and IAPNA\footnote{For evaluation, we consider APP and IAPNA, as we did not find any other implementations of AP algorithms available online. Although the implementation of IAPNA was not readily available, we implemented it in Python for the purpose of our experiments, given its relatively straightforward nature.}) by also discussing the APP scalability benefits. Furthermore, to show the applicability of APP to a real scenario, we consider a diachronic document corpus and we discuss a case-study in the field of computational linguistics and semantic shift detection. In addition, we also provide some evaluation results on a reference benchmark.\\

The article is organised as follows. In Section~\ref{sec:related_work}, the traditional AP algorithm as well as its main, incremental extensions are over-viewed. The APP algorithm is presented in Section~\ref{sec:incremental}. The comparison against benchmark algorithms is discussed in Section~\ref{sec:computational_experiments}. Section~\ref{sec:application} illustrates the application of APP to a case-study of semantic shift detection. Finally, concluding remarks are given in Section~\ref{sec:conclusion}.

\section{Related work}\label{sec:related_work}
Work related to incremental clustering over dynamic datasets and temporal/stream-based data aggregation techniques are widely discussed in the literature (e.g.,~\citealp{aggarwal2018survey,Mansalis2018evaluation,Mei2005Discovering}).
In this context, the APP algorithm we are proposing is conceived as an extension of the original Affinity Propagation algorithm~\citep{frey2007clustering}. For this reason, in the following, we first recall the main features of Affinity Propagation, and then we review the main incremental extensions of this algorithm, by also highlighting the distinctive features of our APP algorithm with respect to the considered solutions.

\subsection{Affinity Propagation}
Affinity Propagation (AP) is a clustering algorithm based on ``message passing" between data points represented as connected nodes on a bipartite graph, in which edges represent the similarity between pairs of points. The main advantages is that, unlike other clustering algorithms such as K-Means or K-Medoids, it does not require the number of clusters to be determined beforehand since they are formed around exemplary nodes, namely \textit{exemplars}, which are representative nodes of the clusters. 
The objective function is to maximise
\begin{equation}
    z = \sum^n_{i=1} s(i, c_i) + \sum^n_{k=1} \delta_k(\mathbf{c})
\end{equation}
where $s(i, c_i)$ denotes similarity between a node $\mathbf{x}_i$ and its nearest exemplar $\mathbf{x}_{c_i}$, and $\delta_k(\mathbf{c})$ has the form
\begin{equation}
\delta_k(\mathbf{c}) = \left\{\begin{matrix}
-\infty& \text{if} \ c_k\neq k \ \text{but} \ \exists i : c_i = k\\ 0 & \text{otherwise} 
\end{matrix}\right.
\end{equation}
and penalises invalid configurations where a node $\mathbf{x}_i$ chooses another nodes $\mathbf{x}_k$ as its exemplar without $\mathbf{x}_k$ being labelled as an exemplar. The optimization problem is implemented by exchanging two kinds of message between nodes on the graph: 
\begin{enumerate}
    \item \textit{responsibility} $r(i, k)$, sent from node $\mathbf{x}_i$ to the candidate exemplar $\mathbf{x}_k$ indicates to what extent $\mathbf{x}_k$ is a good exemplar for $\mathbf{x}_i$.
    \item \textit{availability} $a(i, k)$, sent from the candidate exemplar $\mathbf{x}_k$ to node $\mathbf{x}_i$ indicates to what extent it would be for $\mathbf{x}_i$ to choose $\mathbf{x}_k$ as its exemplar taking into account the accumulated evidence obtained from other nodes about the suitability of  $\mathbf{x}_k$ as an exemplar.
\end{enumerate}
According to~\citep{frey2007clustering}, $r(i, k)$ and $a(i, k)$ can be computed as follows:
\begin{equation}\label{eq:app1}
\resizebox{0.7\hsize}{!}{$
    r(i, k) \leftarrow s(i, k) - \max_{k', \ k'\neq k}\left\{a(i, k') + s(i, k')\right\}
    $}
\end{equation}
\begin{equation}\label{eq:app2}
\resizebox{0.7\hsize}{!}{$
    a(i, k) \leftarrow \min 
    \left\{0, r(k, k) + \sum_{i', \ i' \notin \{i,k\}} \max \left \{0, r(i', k)\right\}
    \right\}
    $}
\end{equation}
Unlike the other pairs, the so called \textit{self-availability} $a(k, k)$ is computed as
\begin{equation}
    a(k, k) = \sum_{i', i'\neq k} \max\left\{0, r(i', k)\right\}.
\end{equation}

In the beginning, all messages are initialised to 0. Then, AP iteratively updates responsibilities and availabilities until convergence. The number of resulting clusters is determined by the clustering algorithm. However, it was argued by~\citeauthor{frey2007clustering} 
that it is influenced by the self-similarity value $s(i, i)$, which is called \textit{preference}, and by the \textit{damping} factor which damps the responsibility and availability of messages to avoid numerical oscillations in the updates. 

As a general remark,~\citeauthor{frey2007clustering} 
suggest preference $p$ should be the median, or minimum value of similarities and point out that a larger $p$ generates a larger number of
clusters
. The damping factor $d$ should be at least 0.5 and less than 1. In particular, the responsibility and availability messages are ``damped'' as follows
\begin{equation}
    \mathbf{msg}_{new} = d\cdot \mathbf{msg}_{old} + (1-d)\cdot \mathbf{msg}_{new}
\end{equation}
where $\mathbf{msg}_{old}$ and $\mathbf{msg}_{new}$ are the values of $a(i, k)$ and $r(i, k)$ before and after the update, respectively.

\subsection{Incremental extensions of Affinity Propagation}
AP was designed for discovering patterns in static data. Several extensions have been proposed to cope with data appearing in a dynamic manner. Incremental extensions of AP have been successfully employed in a series of problems such as text clustering~\citep{shi2009incremental}, robot navigation~\citep{lionel2012unsupervised}, multispectral images classification~\citep{yang2013incremental}, and, recently, semantic shift detection~\citep{periti2022done}. Moreover, we also consider incremental AP extensions where a notion of \textit{clustering history} is somehow supported, that is the capability to trace the object membership over time or to compare clusters related to different time steps. A comparative overview of the considered AP extensions is provided in Table~\ref{tab:sota}.
\begin{table*}[!ht]
\centering
\resizebox{\textwidth}{!}{%
\begin{tabular}{cccccc}
\textbf{Work} &
  \textbf{Learning} &
  \textbf{Basic algorithms} &
  \textbf{Clustering history} &
  \textbf{Efficiency} &
  \textbf{Description} \\ \hline
\multicolumn{1}{|c}{STRAP~\citep{zhang2008frugal}} &
  Unsupervised &
  AP &
  No &
  faster than AP &
  \multicolumn{1}{c|}{\begin{tabular}[c]{@{}c@{}}\\ STRAP assignes new objects to \\ previously generated clusters based \\ on their similarity. \\ \\\end{tabular}} \\ \hline
\multicolumn{1}{|c}{I-APC~\citep{shi2009incremental}} &
  Semi-supervised &
  AP &
  No &
  slower than AP &
  \multicolumn{1}{c|}{\begin{tabular}[c]{@{}c@{}}\\I-APC injects supervision in AP by \\ adjusting the similarity matrix.\\ \\\end{tabular}} \\ \hline
\multicolumn{1}{|c}{ID-AP~\citep{shi2009incremental}} &
  Semi-supervised &
  AP &
  No &
  slower than AP &
  \multicolumn{1}{c|}{\begin{tabular}[c]{@{}c@{}}\\ID-AP injects supervision in AP by \\ adjusting the similarity matrix, and \\ discard useless labeled objects at each\\ time-step.\\\\\end{tabular}} \\ \hline
\multicolumn{1}{|c}{IAPKM~\citep{leilei2014incremental}} &
  Unsupervised &
  \begin{tabular}[c]{@{}c@{}}AP\\ K-Medoids\end{tabular} &
  No &
  faster than AP &
  \multicolumn{1}{c|}{\begin{tabular}[c]{@{}c@{}}\\IAPKM adjusts the current clustering \\ results according to new objects by \\ combining AP and K-Medoids.\\\\\end{tabular}} \\ \hline
\multicolumn{1}{|c}{IAPNA~\citep{leilei2014incremental}} &
  Unsupervised &
  \begin{tabular}[c]{@{}c@{}}AP\\ Nearest Neighbors\end{tabular} &
  No &
  faster than AP &
  \multicolumn{1}{c|}{\begin{tabular}[c]{@{}c@{}}\\In IAPNA, responsibilities and \\ availabilities of the new objects are \\ assigned referring to their Nearest \\ Neighbor among the previous objects.\\\\\end{tabular}} \\ \hline
\multicolumn{1}{|c}{EAP~\citep{arzeno2017evolutionary,arzeno2021evolutionary}} &
  Unsupervised &
  AP &
  Yes &
  faster than AP &
  \multicolumn{1}{c|}{\begin{tabular}[c]{@{}c@{}}\\EAP trace the clustering history by\\ introducing consensus nodes and \\ factors into the AP graph.\\\\\end{tabular}} \\ \hline
\multicolumn{1}{|c}{SED Stream-AP~\citep{sunmood2018evolution}} &
  Unsupervised &
  \begin{tabular}[c]{@{}c@{}}AP\\ SED-Stream\end{tabular} &
  Yes &
  slower than AP &
  \multicolumn{1}{c|}{\begin{tabular}[c]{@{}c@{}}\\SED Stream-AP trace the clustering history\\ by combining the SED-Stream and AP\\ clustering algorithms.\\\\\end{tabular}} \\ \hline
  \multicolumn{1}{|c}{APP} &
  Unsupervised &
  \begin{tabular}[c]{@{}c@{}}AP\end{tabular} &
  Yes &
  faster than AP &
  \multicolumn{1}{c|}{\begin{tabular}[c]{@{}c@{}}\\APP traces the cluster history by\\consolidating past clustering results through\\the use of cluster centroids, and discards \\obsolete objects at each time-step by\\enforcing cluster pruning.\\\end{tabular}} \\ \hline
\end{tabular}%
}
\caption{Summary view of incremental extensions of AP.}
\label{tab:sota}
\end{table*}

\paragraph{STRAP: Streaming AP}
~\cite{zhang2008frugal} 
propose an incremental AP clustering algorithm (STRAP) for data streaming settings that reduces the time complexity of AP by limiting the number of its recomputations.
The idea is to assign new objects to previously generated clusters only if they satisfy a similarity requirement with respect to the current exemplars. On the contrary, a reservoir is leveraged to detain too dissimilar objects. When the size of reservoir exceeds a threshold, or some changes in the rate of acquisition are detected, the AP is re-executed over the current exemplars and the 
objects in the reservoir. An additional step is employed to merge the exemplars independently learned from subsets of the whole dataset.

\paragraph{I-APC: Incremental AP clustering}
\cite{shi2009incremental} 
propose a semi-supervised incremental AP (I-APC) which injects some supervision in the clustering by adjusting the similarity matrix of the AP algorithm. 
They set much larger distance for objects with the same label and much smaller distance for objects with different labels. At each time-step, after each AP run, the labeled dataset is extended with the most similar objects to the current clusters, and the similarity matrix is reset according to the new labeled data. However, this step affects computational time and it makes I-APC cost more CPU time than AP.

\paragraph{ID-AP: Incremental and Decremental Affinity Propagation}
Similarly to~\cite{shi2009incremental}, ~\cite{yang2013incremental} 
propose a semi-supervised incremental algorithm, called Incremental and Decremental AP (ID-AP), that incorporates a small number of labeled samples to guide the clustering process of the conventional AP algorithm. 
At each time-step the labeled samples are used as prior information to adjust the similarity matrix of the AP algorithm. Furthermore, the algorithm deals with the \textit{stability-plasticity} dilemma by using an incremental and a decremental learning approach for selecting the most informative unlabeled data and discarding useless labeled samples, respectively. The intrinsic relationship between the labeled samples and unlabeled data improves the clustering performance. On the other hand, the learning phase of ID-AP method is several times higher than that required from the conventional AP since the selection/discard phases involve repeated execution of the clustering algorithm. 

\paragraph{IAPKM: Incremental Affinity Propagation based on K-Medoids}
\cite{leilei2014incremental} present an Incremental Affinity Propagation based on K-Medoids (IAPKM). The goal of this extension is to adjust the current clustering results according to new incoming objects, rather than recomputing AP clustering on the whole data set. IAPKM combines AP and K-Medoids in an incremental clustering task, that is: AP clustering is executed on the initial bunch of objects, and K-Medoids is employed to modify the current clustering result according to the new arriving objects. As a result, IAPKM achieves comparable clustering performance and can save a great deal of time compared to the conventional AP algorithm. However, the number of clusters cannot be adjusted according to the new incoming objects since the traditional K-Medoids can't adjust the number of clusters automatically.

\paragraph{IAPNA: Incremental Affinity Propagation based on Nearest Neighbor Assignment}
As an alternative to IAP-KM, ~\cite{leilei2014incremental} 
discuss an Incremental version of Affinity Propagation based on Nearest Neighbor Assignment (IAPNA). 
The intuition under IAPNA is that objects added at different time-steps are at different statuses: pre-existing objects have established certain relationships (nonzero responsibilities and nonzero availabilities) between each other after AP, while new objects' relationships with other objects are still at the initial level (zero responsibilities and zero availabilities). The idea of IAPNA is to put all the data points at the same status before proceeding with the AP procedure till convergence. According to this idea, responsibilities and availabilities of the new incoming objects are assigned referring to their nearest neighbors. Similarly to IAPKM, IAPNA achieves higher performance than traditional AP clustering while reducing computational complexity. In addition, it preserves the AP feature of automatically discovering new clusters.

\paragraph{EAP: Evolutionary Affinity Propagation}
An Evolutionary Affinity Propagation (EAP) is presented by
~\cite{arzeno2017evolutionary,arzeno2021evolutionary}. Compared to previous incremental extensions of AP, EAP is the first algorithm that can automatically trace the
clustering history and temporal changes in cluster memberships across time. EAP introduces consensus nodes and factors into the AP graph with the aim to encourage objects to select a consensus node, rather than another object, as their exemplar. Clusters are traced by observing the positions of consensus nodes in the clustering history. Basically, the creation and the disappearance of consensus nodes indicate cluster birth and death, respectively. In EAP, the computational time is also reduced since messages need to be passed between consensus nodes and not between all pairs of objects.

\paragraph{SED Stream-AP: Evolutionary Affinity Propagation}
\cite{sunmood2018evolution} propose the evolutionary clustering SED-Stream-AP as an integration of the SED-Stream~\citep{waiyamai2014sed} and the AP clustering algorithms. 
SED-Stream-AP adopts a two stage process phases, called \textit{online} and \textit{offline} phase,  respectively. 
In the online phase, the clustering history is continuously monitored and detected. The evolution-based clustering of SED-Stream enables SED-Stream-AP to support different evolving structures (e.g., appearance, merge). 
In the off-line phase, the AP clustering is used to automatically determine the number of clusters and deliver the final clustering without any need for user intervention.\\

\subsubsection{Framing APP with respect to the above solutions} 
Inspired by STRAP~\citep{zhang2008frugal}, the APP algorithm we propose performs clustering over exemplars created in past aggregation stages and new incoming objects. As a difference with STRAP, APP ignores cluster exemplars and replaces them by using an average representation of the clusters, i.e. the centroid of each cluster. Moreover, new incoming objects are {\em a posteriori} clustered and not {\em a priori} assigned to a previously generated cluster. In particular,  APP replaces the use of a reservoir with the assumption of ``group evolution'', meaning that a new cluster for a new kind of objects can be detected only if there is a relevant number of incoming exemplar objects associated with it. 

Similarly to ID-AP~\citep{yang2013incremental}, APP is an incremental extension of AP conceived for dealing with the stability-plasticity dilemma by enforcing faithfulness and forgetfulness in evolutionary scenarios. Like  SED-Stream-AP~\citep{sunmood2018evolution}, APP can trace the clustering history by supporting different kinds of cluster stratifications. 

\section{A-Posteriori affinity Propagation}\label{sec:incremental}
Using the conventional AP algorithm to cluster dynamic datasets is not suitable to cope with the stability-plasticity dilemma~\citep{yang2013incremental}. In particular, clusters generated at time $t-1$ can be mixed-up due to a new bunch of objects that arrive at time $t$. This means that previously clustered objects at time $t-1$ can remain in the same cluster at time $t$, but they can also be moved to another cluster due to the updated object picture from time $t-1$ to time $t$. In this situation, tracing the history of a specific cluster across different time periods becomes arduous, and a number of noisy clusters could be created when different kinds of objects arrive according to a skewed distribution~\citep{martinc2020capturing}.

\begin{figure*}[!ht]
\centering
\includegraphics[width=0.75\textwidth]{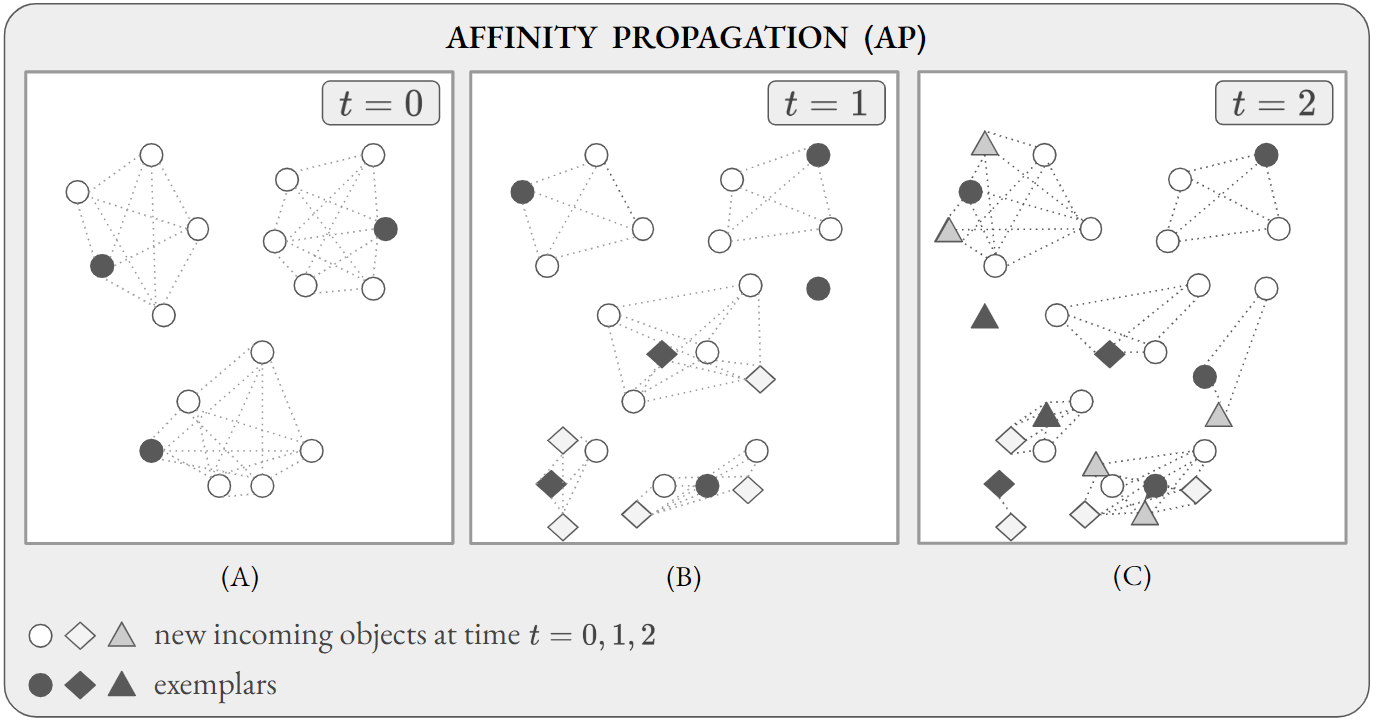}
\caption{Example of AP with an incremental scenario. (A) shows the clustering result over the initial bunch of objects ($t=0$) represented by white circles. The black objects denote the cluster exemplars and dashed lines connect the objects of each cluster. (B) show the the clustering result after the second AP run ($t=1$). New incoming objects at time $t=1$ are represented by gray diamonds. Similarly to (B), the clustering result after the third AP run ($t=2$) is shown in (C). New incoming objects at time $t=2$ are represented by gray triangles.}
\label{fig:ap}
\end{figure*}

Figure~\ref{fig:ap} shows an example of AP clustering illustrating such a problem. The conventional AP clustering is implemented on the initial bunch of objects ($t=0$), represented by white circles. The clustering result is shown in Figure~\ref{fig:ap}A, where the black objects denote the cluster exemplars. The new objects represented by gray diamonds and triangles  arrive at time $t=1$ and $t=2$, respectively. After the arrival of new objects, the clustering result of the second and third AP run is shown in Figure~\ref{fig:ap}B-C. By comparing Figure~\ref{fig:ap}A-B-C, we note that some objects change cluster in the various AP rounds and several clusters are generated ($t=2$).

In the following, we present APP. The objects to cluster become progressively available at different time-steps $t = \{ 0, \dots, n \}$. At each time-step $t$, APP clusters the new incoming objects \textit{a-posteriori} by considering a consolidated version of the clusters created at time $t-1$. For each cluster, the AP notion of exemplar is replaced by {\em centroid} and it is defined as a summary representation of the associated objects with the aim to consolidate the cluster observed until $t-1$. In particular, in the remainder of the paper, we work with objects that are data points, namely vectors of numerical features. In this context, a cluster centroid is computed as average representation of the associated object vectors. As a main difference with AP, in APP, the objects previously clustered do not change cluster when new objects arrive and clusters generated in a certain time-step are
consolidated/stratified over the past ones.



\subsection{The APP algorithm}
Algorithm~\ref{alg:app} provides the pseudo-code of the proposed APP. 
\begin{algorithm}
{\small 
\caption{\textit{The APP algorithm}}\label{alg:app}
 \hspace*{\algorithmicindent} \textbf{Input} \\
\hspace*{\algorithmicindent} $t$: \textit{time-step}\\
\hspace*{\algorithmicindent} $X$: \textit{objects at time-step} $t$\\
\hspace*{\algorithmicindent} $X_1$: \textit{objects at time-step} $t-1$\\
\hspace*{\algorithmicindent} $L_1$: \textit{labels at time-step} $t-1$\\
\hspace*{\algorithmicindent} $th_{\gamma}$: \textit{pruning threshold} \\\\
\hspace*{\algorithmicindent} \textbf{Output} \\
\hspace*{\algorithmicindent} $L, X$: \textit{at time-step} $t$\\
 \begin{algorithmic}[1]
\If{t == 0}
    \State L $\gets$ $AP$(X)
    \\
\Else
    \State $\mu$X$_1$ $\gets$ $Pack$(L$_1$, X$_1$)
    \State L$_2$ $\gets$ $AP$( $\mu$X$_1$ $\cup$ X )
    \State $\mu$L$_1$, L $\gets$ \textit{Split}(L$_2$)
    \State L$_1$ $\gets$ $UnpackAndUpdate$($\mu$L$_1$, $\mu$X$_1$, L$_1$, X$_1$)
    \State L, X $\gets$ $Pruning$($ \ $L$_1$ $\cup$ L, $ \ $ X$_1$ $\cup$ X, $ \ $  $th_{\gamma}$)  
\EndIf\\
\State \textbf{yield} L, X
\end{algorithmic}}
\end{algorithm}
Let's call $X$ and $X_1$, and $L$ and $L_1$ the objects and the cluster labels at time $t$ and $t-1$, respectively. At time $t=0$, the execution of APP coincides with the conventional AP algorithm. At each time $t>0$, for each existing cluster computed at time $t-1$, the objects $x_i \in X_1$ are packed into a single representation called cluster centroid $\mu$. The set of the centroids for $X_1$ is denoted $\mu X_1$. Then, the conventional AP algorithm is executed on $\mu X_1 \cup X$, with the aim to obtain a new set of temporary labels $L_2$, i.e., the new assignment of objects to clusters. Such labels are then split in two subsets, $\mu L_1$ and $L$, which contain labels for each average representation in $\mu X_1$ and for each object in $X$, respectively. Given $\mu L_1, \mu X_1, L_1, X_1$, APP unpacks the centroids of $\mu L_1$ into the corresponding objects $X_1$ mapping the previous labels $L_1$ into the new labels of their respective centroids $\mu L_1$. Finally, APP returns $ L_1 \cup L$, which is the union of the unpacked and updated $L_1$ and $L$. 

The APP algorithm enforces \textit{faithfulness} and \textit{forgetfulness} as described in the following.\\

\noindent {\bf Faithfulness} is the capability to preserve clustering history possibly enriched with new objects. At time $t$, the execution of APP ensures that the objects $X_1$ arrived in previous time-steps do not change cluster. Indeed, each cluster existing at time $t-1$ is summarised by a centroid defined as an average representation of the cluster objects associated with it through $L_1$. The centroids are not changed by the APP execution at time $t$, thus also the objects arrived until $t-1$ cannot change cluster. As a result, the clusters of time $t-1$ and the associated centroids constitute the ``memory'' of the objects observed in the past. In APP, the centroids of clusters at time $t-1$ are exploited as additional objects to cluster together with the new incoming objects at time $t > 0$. The new objects are stratified over the existing clusters according to one of the following criteria:

\begin{itemize}
    \item \textit{stratification-by-creation}: a new cluster is created containing a subset of the new incoming objects $\bar{X} \subseteq X$ when all the objects in $\bar{X}$ are found to be too dissimilar from all the existing cluster centroids $\mu X_1$.
    
    \item \textit{stratification-by-enrichment}: a previously created cluster is enriched with a subset of the new incoming objects $\bar{X} \subseteq X$ when all the objects in $\bar{X}$ are found to be similar to a cluster centroid in $\mu X_1$.
    
    \item \textit{stratification-by-merge}: a new, unique cluster is created by merging two or more centroids in $\mu X_1$ and a subset of the new incoming objects $\bar{X} \subseteq X$ when the objects in $\bar{X}$ are found to be similar to all the merged centroids. 
    
\end{itemize}

\noindent {\bf Forgetfulness} is the capability to recognise obsolete clusters and discard them. At a certain time $t$, it is possible that a cluster represents the memory of a group of {\em obsolete objects}, namely a group emerged in past time-steps, but disappeared in recent observations. To enforce forgetfulness, APP allows to drop the clusters that represent obsolete groups of objects. Each cluster is associated with an {\em aging index} $\gamma \leq t$ that denotes the last time-step $t$ in which the cluster has been created/changed. For instance, a cluster enriched by new objects at time $t$ has an aging index $\gamma = t$. A {\em pruning threshold} $th_{\gamma} \in [1, \ +\infty]$ is defined in APP to define when a cluster can be considered obsolete. The threshold specifies the max number of APP rounds that can be executed without any change on a cluster contents. At each time-step, each cluster defined by $L$ is evaluated for possible pruning with respect to $th_{\gamma}$. Given a cluster with aging index $\gamma$, the cluster is pruned when $t-\gamma > th_{\gamma}$. When $th_{\gamma} \geq t$, it means that forgetfulness is not enforced and all the clusters created at any time-step is maintained. Otherwise, forgetfulness is enforced and the pruning condition is applied. For instance when $th_{\gamma} = 1$ and $th_{\gamma} < t$, all the clusters not enriched at the last time $t$ are considered as obsolete, and then pruned. 

\begin{figure*}[!ht]
\centering
\includegraphics[width=\textwidth]{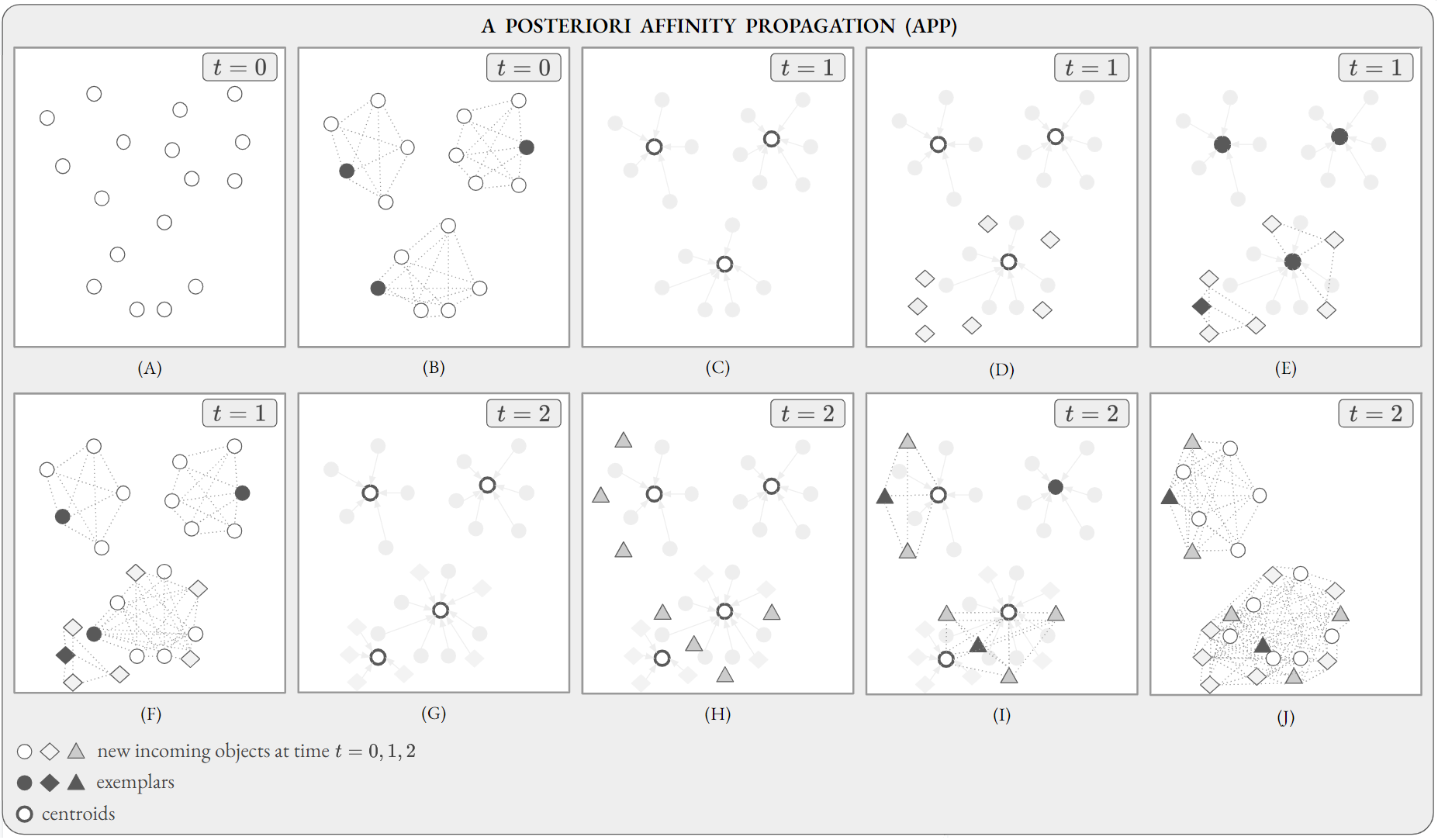}
\caption{Example of APP. (A) shows the objects available at time $t = 0$. The first clustering result coincides with AP and it is represented in (B). The black objects denote the cluster exemplars. For the sake of clarity, dashed lines fully connect the objects of each cluster. (C) shows the cluster centroids as bold circles generated by averaging the objects of each cluster on the background. (D) shows the input objects of APP at time $t=1$. Gray diamonds represent the new incoming objects. The clustering result is represented in (E). In (F), cluster centroids are unpacked and their cluster labels are associated with each object they previously packed. The second APP run at time $t=2$ is shown in (G)-(H)-(J). New incoming objects are represented by gray triangles. (J) denotes the final clustering result. Note that the cluster on the right-top corner of (I) disappears in (J) due to a pruning threshold $th_{\gamma}=1$.}
\label{fig:app}
\end{figure*}

Figure~\ref{fig:app} is an example of APP execution with pruning threshold $th_{\gamma} = 1$. The initial bunch of objects ($t=0$) is shown in Figure~\ref{fig:app}A. The clustering result at time $t=0$ is represented in Figure~\ref{fig:app}B. Black objects denote the cluster exemplars.
In Figure~\ref{fig:app}C, centroids are calculated as average representations of cluster objects ($t=1$) and they are denoted as bold circles.
New objects at time ($t=1$) are represented as gray diamonds in Figure~\ref{fig:app}D. After the cluster consolidation, the clustering result of the APP run is shown in Figure~\ref{fig:app}E ($t=1$). In particular, Figure~\ref{fig:app}E shows an example of stratification-by-creation (i.e., cluster on the bottom-left corner) and an example of stratification-by-enrichment (i.e., cluster on the bottom-middle part). In Figure~\ref{fig:app}F, each centroid is unpacked and its cluster label is associated to each object it had previously packed. The consecutive round of APP ($t=2$) is presented in Figure~\ref{fig:app}G-H-J. In particular, Figure~\ref{fig:app}I shows an example of stratification-by-merge where two previously generated clusters are merged into a single one.
The final clustering result at time $t=2$ is shown in Figure~\ref{fig:app}J. As a result of the stratification-by-pruning, the cluster on the right-top corner in Figure~\ref{fig:app}I is pruned in Figure~\ref{fig:app}J since it is unchanged for two iterations. As a difference with AP (see Figure~\ref{fig:ap}), objects do not change cluster in Figure~\ref{fig:app} and a lower number of clusters is generated.

\subsection{Complexity and Memory Usage Analysis}
Since APP leverages AP for object clustering, the complexity of APP and AP are related. In AP, the time complexity of message-passing iteration according to Equations~\ref{eq:app1} and~\ref{eq:app2} is $\mathcal{O}(N^2)$, where $N$ is the number of all the current available objects. Therefore, the time complexity is $\mathcal{O}(N^2T)$, where $T$ is the number of iterations until convergence. Further, the memory complexity is in the order $\mathcal{O}(N^2)$ if a dense similarity matrix is used. 

Similarly, the time complexity of APP is $\mathcal{O}(M^2T_1)$, where $M = (\mu_{t-1} + n_t)$, and $\mu_{t-1}$, $n_t$ are the number of previous centroids and the number of the new incoming objects, respectively. At each iteration, the memory complexity of APP is $\mathcal{O}(M^2)$, in that, there is no need to keep in memory previously clustered objects during the AP execution of APP (Algorithm~\ref{alg:app}, row 6). By definition $M << N$ and $T_1 << T$, thus a lot of time and memory are saved, making APP a scalable solution in incremental scenarios. 
Moreover, when $th_{\gamma} > 0$, time and memory complexity are further reduced to $\mathcal{O}(M_\gamma^2T_2)$, $\mathcal{O}(N_\gamma)$, respectively; where $M\gamma = (\mu_{t-1}^{(\gamma)} + n_t)$ and $\mu_{t-1}^{(\gamma)}$ is the number of previous centroids that were not affected by pruning, and $T_2 < T_1$. Basically, the smaller $\gamma$, the more $\mu_{t-1}^{(\gamma)} < \mu_{t-1}$, since more clusters will be pruned.

\section{Experiments and evaluation results}\label{sec:computational_experiments}
The goal of our experimentation is to compare the results of APP against benchmark clustering algorithms. For comparison with APP, we selected AP since it is the baseline clustering algorithm on which APP relies upon, and IAPNA since it is a well-known and top-cited incremental extension of AP. In the evaluation, we first focus on two evaluation experiments called \textit{uniform-incremental} and \textit{variable-incremental} experiments. Both the experiments are based on a dynamic scenario where the objects to cluster arrive as separated bunches at different time-steps. In the uniform-incremental experiment, we define the number and the set of objects arriving at the various time-steps without any constraint on the category. The idea is to analyse the behavior of the considered clustering algorithms on a pure incremental setting like the one proposed in~\citep{leilei2014incremental} (see Section~\ref{sec:incremental_scenario}).
In the variable-incremental experiment, the category of the objects arriving at each time-step is constrained according to a given schema. The idea is to analyse the capability of the considered clustering algorithms to recognise the categories of the incoming objects when they appear over time according to a specific incremental schema, that can be growing, shrinking, or stable (see Section~\ref{sec:evolutionary_scenario}). 

All the experiments are implemented in Python 3.10 and they are conducted on a PC with 1.80GHz Intel Core i7 processor and 16GB of RAM. Our code is based on the implementation of AP by scikit-learn\footnote{ \url{scikit-learn.org/stable/}}. The APP code is available at \url{https://github.com/umilISLab/APP}.

\subsection{Datasets and pre-processing}
In the evaluation, four popular labeled datasets are considered. In particular, we selected Iris, Wine, and Car datasets from~\citep{newman1998uci} since they are used in the evaluation of AP and IAPNA 
by~\cite{leilei2014incremental}. Moreover,  we added the KDD-CUP dataset since it is characterised by a high number of categories~\citep{sunmood2018evolution}, and thus it is appropriate for clustering evaluation in incremental experiments. In all the datasets, the objects are described as feature vectors; a different number of features per object is defined for each dataset.

A summary view of the benchmark datasets used in the evaluation is provided in Table~\ref{tab:datasets}. 
\begin{table}[!ht]
\centering
\resizebox{0.8\columnwidth}{!}{%
\begin{tabular}{l|cccc}
\multicolumn{1}{c|}{Dataset} &
  \begin{tabular}[c]{@{}c@{}}Number of\\ objects\end{tabular} &
  \begin{tabular}[c]{@{}c@{}}Number of\\ features\end{tabular} &
  \begin{tabular}[c]{@{}c@{}}Number of\\ categories\end{tabular} &
  \begin{tabular}[c]{@{}c@{}}Usage of \\ dataset\end{tabular} \\ \hline
Iris  & 150  & 4  & 3  & whole  \\
Wine  & 178  & 13 & 3  & whole  \\
Car   & 260 & 6  & 4  & partly \\
KDD-CUP & 2904 & 41 & 11 & partly
\end{tabular}%
}
\caption{A summary description of the benchmark datasets.}
\label{tab:datasets}
\end{table}
Some datasets (Car and KDD-CUP) are characterised by a highly unbalanced number of objects per category. As in~\cite{leilei2014incremental}, we select and use only part of them. In particular, we consider 65 objects taken from the top 4 most numerous categories in the Car dataset, and  264 objects taken from the top 11 most numerous categories in the KDD-CUP dataset.

A pre-processing stage is enforced to normalise the dataset objects. 
Since the experiments are performed in a dynamic scenario, a single normalisation stage on the whole dataset is not appropriate. Instead, at each time-step of the experiments, we perform normalisation on the $N_t$ objects of the dataset available at time $t$ 
. For the sake of comparison, we use the same normalisation used by
~\cite{leilei2014incremental}.

\subsection{Evaluation metrics}
As in~\cite{leilei2014incremental}, for clustering objects, we calculate the similarity between pairs of objects through the negative euclidean distance where we do not leverage the preference coefficients described 
by~\cite{leilei2014incremental}. For each dataset, the preference $p$ (self-similarity) is set to the median of the input similarities at a given time (see Section~\ref{sec:related_work} for further details about the $p$ parameter).

The clustering results are evaluated according to \textit{Purity} (PUR) and \textit{Normalised Mutual Information} (NMI). To compute PUR, each cluster is assigned to the category which is most frequent in the cluster, and then the accuracy of this assignment is measured by counting the number of correctly assigned objects and by dividing by $N_t$, that is the number of objects of the dataset available at time $t$. Formally:
\begin{equation}
    PUR(\Omega, \mathcal{C}) = \frac{1}{N_t} \sum_k \max_j \ \bar{\omega}_k \cap \bar{c}_j \ ,
\end{equation}
where $\Omega = \{\omega_1, ..., \omega_K\}$ is the set of clusters, $\mathcal{C} = \{c, ..., c_J\}$ is the set of categories, and $\bar{\omega}_k$ and $\bar{c}_j$ are the set of objects in $\omega_k$ and $c_j$, respectively. High PUR values are frequently achieved when a high number of clusters is generated. For instance, PUR is 1 when each object is placed in a corresponding singleton cluster. Thus, we also exploit NMI to estimate the quality of the clustering by considering the number of generated clusters. NMI is defined as:
\begin{equation}
    NMI(\Omega, \mathcal{C}) = \frac{I(\Omega, \mathcal{C})}{[H(\Omega) + H(\mathcal{C})]/2} \ ,
\end{equation}
where $I(\Omega, \mathcal{C})$ is the mutual information between the set of clusters $\Omega$ and the set of categories $\mathcal{C}$, and the normalisation $[H(\Omega) + H(\mathcal{C})]/2$ is introduced to penalise large cardinalities of $\Omega$ with respect to $\mathcal{C}$, in that, the entropy $H(\Omega)$ tends to increase with the number of clusters.

As in~\cite{leilei2014incremental}, three metrics are employed to evaluate the scalability of the considered clustering algorithms, namely the \textit{Number of Iterations} until convergence (NI), the \textit{Computation Time} (CT) in seconds, and the \textit{Memory Usage} (MU) in MB. Furthermore, we also consider the \textit{Number of Clusters} (NC) generated at each time-step. 

\subsection{Experimental setup}\label{sec:setup}
The setup of uniform-incremental and variable-incremental experiments is discussed in the following. 

As a general remark, we stress that the experiments are repeated 100 times for each dataset; each time, the order of incoming objects is randomly defined. For each dataset, the settings of the 100 executions are stored and used for each considered algorithm (i.e., AP, IAPNA, and APP). We analyse the results by considering the median score of the 100 obtained values at each time-step. 

The hyperparameters of the AP algorithm are configured as follows: the maximum number of iterations is set to 200, the damping factor is set to 0.9, and 15 iterations without changes in the exemplars at the last time-step are required before declaring convergence. 

About IAPNA, since the implementation used in the evaluation of~\cite{leilei2014incremental} is not available, we developed a Python IAPNA implementation for the sake of our experiments.

About the APP configuration, we define a pruning threshold $th_\gamma = 1$\footnote{As pruning threshold, we chose the value that provided the best trade-off between APP performance and scalability in all the considered experiments.}. 

\subsubsection{Uniform-incremental setting}\label{sec:evolutionary_scenario}
In the uniform-incremental setting, we borrow the evaluation setup proposed 
by~\cite{leilei2014incremental}. A fixed (i.e., uniform) number of objects is scheduled for arrival in any time-step without considering the category. Each dataset is shuffled and split through sampling into six bunches (one for each time-step). For each dataset, we define i) the number of incoming objects at the first time-step ($t =0$), and ii) the number of incoming objects at any subsequent time-steps ($t>0$).
In this experiment, most of the objects become available at time-step $0$-th, while few objects are introduced in the subsequent time-steps. The details about dataset sampling in the incremental setting are provided in Table~\ref{tab:incremental}. For instance, considering the IRIS dataset, 100 objects are sampled for clustering at the first time-step, and 10 by 10 objects are sampled in the subsequent time-steps.   
\begin{table}[!ht]
\centering
\resizebox{0.8\columnwidth}{!}{%
\begin{tabular}{l|cc}
\multicolumn{1}{c|}{Dataset} &
  \begin{tabular}[c]{@{}c@{}}Number of objects\\ (first time-step)\\ objects\end{tabular} &
  \begin{tabular}[c]{@{}c@{}}Number of objects\\ (subsequent time-steps)\end{tabular} \\ \hline
Iris  & 100 & 10 \\
Wine  & 128 & 10 \\
Car   & 210 & 10 \\
KDD-CUP & 1904 & 200
\end{tabular}%
}
\caption{The number of objects in the incremental setting (first and subsequent time-steps).}
\label{tab:incremental}
\end{table}

\subsubsection{Variable-incremental setting}\label{sec:incremental_scenario}
In the variable-incremental experiment, the number of incoming objects at each time-step is not fixed/uniform. The goal is to analyse the behavior of clustering algorithms when a larger number of incoming objects is scheduled for arrival at each time-step with respect to the uniform-incremental experiment. Moreover, the category of the objects arriving at each time-step is chosen according to a specific incremental schema. Each dataset is shuffled and split through sampling into six bunches (one for each time-step). The object sampling from each category in a given time-step is defined according to one of the following schema/behavior: 
\begin{enumerate}
    \item \textit{growing}, the objects of a category are sampled by scheduling the order of arrival to be ascending in size across the time-steps. The category reproduces the behavior of a growing group of objects over time.
    \item \textit{shrinking}, the objects of a category are sampled by scheduling the order of arrival to be decreasing in size across the time-steps. The category reproduces the behavior of a shrinking group of objects over time.
    \item \textit{stable}, an equal number of objects of a category is scheduled for arrival in any time-step. The category reproduces the behavior of a stable group of object over time.
\end{enumerate}
In each of the 100 iterations, each category of the datasets is associated with a certain schema with a $33\%$ probability (i.e., the three schemas are equally probable over the categories). The arrival of objects of growing and shrinking categories can be focused in a subset of the time-steps. This means that the objects of a growing category can start to appear in a time-step $t > 0$, as well as the objects of a shrinking category can be consumed before the last time-step. As a consequence, in a given time-step, the objects of a category can be missing. Otherwise, according to the ``group evolution'' assumption, a minimum number of object $q$ of a category is scheduled for arrival in any time-step $t$ according to the associated schema. The aim is that any category appearing in a certain time-step has enough objects for being recognised by the clustering algorithms. As a final constraint, we define that the incoming objects at each time-step are taken from two different categories as a minimum. 

In the experiment, for each category, we define $q$ as the $10\%$ of the dataset size divided by the number of dataset categories. A summary of $q$ values for the categories of each dataset is provided in Table~\ref{tab:qparam}. 
\begin{table}[!ht]
\centering
\resizebox{0.35\columnwidth}{!}{%
\begin{tabular}{l|c}
\multicolumn{1}{c|}{Dataset} &
  \begin{tabular}[c]{@{}c@{}}$q$\\ parameter\end{tabular} \\ \hline
Iris  & 5\\
Wine  & 6\\
Car   & 7\\
KDD-CUP & 26 
\end{tabular}%
}
\caption{The minimum number of objects $q$ per dataset category in the variable-incremental setting.}
\label{tab:qparam}
\end{table}

\subsection{Experimental results}
All the considered algorithms (i.e., AP, IAPNA, and APP) are based on AP for clustering objects in the first time-step. Thus, the results of the three algorithms coincide on the first clustering execution at time $t=0$. For this reason, the results on the $0$-th bunch of objects are not shown/considered in the analysis. 

\subsubsection{Results on the uniform-incremental experiment}
Experimental results with the uniform-incremental settings are shown in Tables~\ref{tab:comparison_pur},~\ref{tab:comparison_nmi},~\ref{tab:comparison_ct},~\ref{tab:comparison_mu},~\ref{tab:comparison_ni},~\ref{tab:comparison_nc}. 

\begin{table}[!ht]
\centering
\resizebox{0.90\columnwidth}{!}{%
\begin{tabular}{l|c|cllll}
\multicolumn{1}{c|}{Dataset} &
  Method & 1th & 2th & 3th & 4th & 5th \\ \hline
Iris &
  \begin{tabular}[c]{@{}c@{}}AP\\ IAPNA\\ APP\end{tabular} &
  \begin{tabular}[c]{@{}c@{}}0.964*\\0.882\\ \textbf{0.873}\\\end{tabular} &
  \begin{tabular}[c]{@{}l@{}}0.975*\\0.950\\ \textbf{0.867}\end{tabular} &
  \begin{tabular}[c]{@{}l@{}}0.954*\\ 0.877\\ \textbf{0.862}\end{tabular} &
  \begin{tabular}[c]{@{}l@{}}0.957*\\ 0.957*\\ \textbf{0.864}\end{tabular} &
  \begin{tabular}[c]{@{}l@{}}0.967*\\ 0.953\\ \textbf{0.667}\end{tabular} \\ \hline
Wine &
  \begin{tabular}[c]{@{}c@{}}AP\\ IAPNA\\ APP\end{tabular} &
  \begin{tabular}[c]{@{}c@{}}0.754\\0.884*\\\textbf{0.710}\end{tabular} &
  \begin{tabular}[c]{@{}l@{}}0.750*\\0.365\\\textbf{0.655}\end{tabular} &
  \begin{tabular}[c]{@{}l@{}}0.747*\\0.620\\\textbf{0.665}\end{tabular} &
  \begin{tabular}[c]{@{}l@{}}0.732*\\0.613\\\textbf{0.661}\end{tabular} &
  \begin{tabular}[c]{@{}l@{}}0.730*\\0.624\\\textbf{0.663}\end{tabular} \\ \hline
Car &
  \begin{tabular}[c]{@{}c@{}}AP\\ IAPNA\\ APP\end{tabular} &
  \begin{tabular}[c]{@{}c@{}}0.814*\\0.791\\\textbf{0.727}\end{tabular} &
  \begin{tabular}[c]{@{}l@{}}0.830*\\0.796\\\textbf{0.604}\end{tabular} &
  \begin{tabular}[c]{@{}l@{}}0.812*\\0.804\\\textbf{0.704}\end{tabular} &
  \begin{tabular}[c]{@{}l@{}}0.816\\0.828*\\\textbf{0.514}\end{tabular} &
  \begin{tabular}[c]{@{}l@{}}0.812\\0.823*\\\textbf{0.550}\end{tabular} \\ \hline
KDD-CUP &
  \begin{tabular}[c]{@{}c@{}}AP\\ IAPNA\\ APP\end{tabular} &
  \begin{tabular}[c]{@{}c@{}}0.863\\0.349\\\textbf{0.816}\end{tabular} &
  \begin{tabular}[c]{@{}l@{}} 0.812*\\0.515\\\textbf{0.806}\end{tabular} &
  \begin{tabular}[c]{@{}l@{}}0.853\\0.512\\\textbf{0.780}\end{tabular} &
  \begin{tabular}[c]{@{}l@{}}0.858\\0.983*\\\textbf{0.741}\end{tabular} &
  \begin{tabular}[c]{@{}l@{}}0.862\\ 0.981*\\\textbf{0.748}\end{tabular} 
\end{tabular}%
}
\caption{Uniform-incremental experiment: comparison on Purity. The highest score is denoted with an asterisk; the APP score is denoted in bold.}
\label{tab:comparison_pur}
\end{table}

\begin{table}[!ht]
\centering
\resizebox{0.90\columnwidth}{!}{%
\begin{tabular}{l|c|cllll}
\multicolumn{1}{c|}{Dataset} &
  Method & 1th & 2th & 3th & 4th & 5th \\ \hline
Iris &
  \begin{tabular}[c]{@{}c@{}}AP\\ IAPNA\\ APP\end{tabular} &
  \begin{tabular}[c]{@{}c@{}}0.600\\ 0.616\\ \textbf{0.707}*\end{tabular} &
  \begin{tabular}[c]{@{}l@{}}0.660\\ 0.658\\  \textbf{0.740}*\end{tabular} &
  \begin{tabular}[c]{@{}l@{}}0.586\\ 0.658\\  \textbf{0.712}*\end{tabular} &
  \begin{tabular}[c]{@{}l@{}}0.561\\ 0.648\\  \textbf{0.718}*\end{tabular} &
  \begin{tabular}[c]{@{}l@{}}0.568\\ 0.594\\  \textbf{0.734}*\end{tabular} \\ \hline
Wine &
  \begin{tabular}[c]{@{}c@{}}AP\\ IAPNA\\ APP\end{tabular} &
  \begin{tabular}[c]{@{}c@{}}0.346\\ 0.582*\\ \textbf{0.363}\end{tabular} &
  \begin{tabular}[c]{@{}l@{}} 0.339\\ 0.000\\ \textbf{0.444}*\end{tabular} &
  \begin{tabular}[c]{@{}l@{}} 0.335\\ 0.484*\\ \textbf{0.444}\end{tabular} &
  \begin{tabular}[c]{@{}l@{}} 0.329\\ 0.489*\\ \textbf{0.445}\end{tabular} &
  \begin{tabular}[c]{@{}l@{}} 0.326\\ 0.565*\\ \textbf{0.417}\end{tabular} \\ \hline
Car &
  \begin{tabular}[c]{@{}c@{}}AP\\ IAPNA\\ APP\end{tabular} &
  \begin{tabular}[c]{@{}c@{}}0.427\\ 0.415\\  \textbf{0.466}*\end{tabular} &
  \begin{tabular}[c]{@{}l@{}}0.432*\\ 0.409\\ \textbf{0.391}\end{tabular} &
  \begin{tabular}[c]{@{}l@{}}0.417*\\ 0.403\\ \textbf{0.221}\end{tabular} &
  \begin{tabular}[c]{@{}l@{}}0.403\\ 0.406*\\ \textbf{0.236}\end{tabular} &
  \begin{tabular}[c]{@{}l@{}}0.392\\ 0.406*\\ \textbf{0.362}\end{tabular} \\ \hline
KDD-CUP &
  \begin{tabular}[c]{@{}c@{}}AP\\ IAPNA\\ APP\end{tabular} &
  \begin{tabular}[c]{@{}c@{}}0.713\\ 0.564\\ \textbf{0.739}*\end{tabular} &
  \begin{tabular}[c]{@{}l@{}}0.700\\ 0.668\\ \textbf{0.743}*\end{tabular} &
  \begin{tabular}[c]{@{}l@{}}0.696\\ 0.665\\ \textbf{0.738}*\end{tabular} &
  \begin{tabular}[c]{@{}l@{}}0.693\\ 0.754*\\ \textbf{0.719}\end{tabular} &
  \begin{tabular}[c]{@{}l@{}} 0.692\\ 0.743*\\ \textbf{0.714}\end{tabular}
\end{tabular}%
}
\caption{Uniform-incremental experiment: comparison on Normalised Mutual Information. The highest score is denoted with an asterisk; the APP score is denoted in bold.}
\label{tab:comparison_nmi}
\end{table}

\begin{table}[!ht]
\centering
\resizebox{0.90\columnwidth}{!}{%
\begin{tabular}{l|c|cllll}
\multicolumn{1}{c|}{Dataset} &
  Method & 1th & 2th & 3th & 4th & 5th \\ \hline
Iris &
  \begin{tabular}[c]{@{}c@{}}AP\\ IAPNA\\ APP\end{tabular} &
  \begin{tabular}[c]{@{}c@{}}0.128\\ 0.241\\ \textbf{0.009}*\end{tabular} &
  \begin{tabular}[c]{@{}l@{}}0.117\\0.221\\ \textbf{0.008}*\end{tabular} &
  \begin{tabular}[c]{@{}l@{}}0.319\\ 0.131\\ \textbf{0.010}*\end{tabular} &
  \begin{tabular}[c]{@{}l@{}}0.321\\ 0.260\\ \textbf{0.009}*\end{tabular} &
  \begin{tabular}[c]{@{}l@{}}0.156\\ 0.238\\ \textbf{0.008}*\end{tabular} \\ \hline
Wine &
  \begin{tabular}[c]{@{}c@{}}AP\\ IAPNA\\ APP\end{tabular} &
  \begin{tabular}[c]{@{}c@{}}0.199\\0.184\\\textbf{0.052}*\end{tabular} &
  \begin{tabular}[c]{@{}l@{}}0.182\\0.123\\\textbf{0.047}*\end{tabular} &
  \begin{tabular}[c]{@{}l@{}}0.204\\0.117\\\textbf{0.051}*\end{tabular} &
  \begin{tabular}[c]{@{}l@{}}0.221\\0.153\\\textbf{0.050}*\end{tabular} &
  \begin{tabular}[c]{@{}l@{}}0.278\\0.364\\\textbf{0.051}*\end{tabular} \\ \hline
Car &
  \begin{tabular}[c]{@{}c@{}}AP\\ IAPNA\\ APP\end{tabular} &
  \begin{tabular}[c]{@{}c@{}}0.332\\0.200\\\textbf{0.074}*\end{tabular} &
  \begin{tabular}[c]{@{}l@{}}0.406\\ 0.678\\\textbf{0.058}*\end{tabular} &
  \begin{tabular}[c]{@{}l@{}}0.563\\0.282\\\textbf{0.028}*\end{tabular} &
  \begin{tabular}[c]{@{}l@{}}0.842\\0.844\\\textbf{ 0.048}*\end{tabular} &
  \begin{tabular}[c]{@{}l@{}}0.867\\0.231\\\textbf{0.035}*\end{tabular} \\ \hline
KDD-CUP &
  \begin{tabular}[c]{@{}c@{}}AP\\ IAPNA\\ APP\end{tabular} &
  \begin{tabular}[c]{@{}c@{}}18.523\\44.656\\\textbf{0.294}*\\\end{tabular} &
  \begin{tabular}[c]{@{}l@{}}26.752\\43.041\\\textbf{0.210}*\end{tabular} &
  \begin{tabular}[c]{@{}l@{}}34.037\\36.304\\\textbf{0.209}*\end{tabular} &
  \begin{tabular}[c]{@{}l@{}}42.068\\83.318\\\textbf{0.211}*\end{tabular} &
  \begin{tabular}[c]{@{}l@{}}46.151\\68.759\\\textbf{0.192}*\end{tabular}
\end{tabular}%
}
\caption{Uniform-incremental experiment: comparison on Computation Time. The highest score is denoted with an asterisk; the APP score is denoted in bold.}
\label{tab:comparison_ct}
\end{table}

\begin{table}[!ht]
\centering
\resizebox{0.90\columnwidth}{!}{%
\begin{tabular}{l|c|cllll}
\multicolumn{1}{c|}{Dataset} &
  Method & 1th & 2th & 3th & 4th & 5th \\ \hline
Iris &
  \begin{tabular}[c]{@{}c@{}}AP\\ IAPNA\\ APP\end{tabular} &
  \begin{tabular}[c]{@{}c@{}}0.303\\0.308\\\textbf{0.020}*\end{tabular} &
  \begin{tabular}[c]{@{}l@{}}0.359\\0.366\\\textbf{0.023}*\end{tabular} &
  \begin{tabular}[c]{@{}l@{}}0.420\\0.428\\\textbf{0.024}*\end{tabular} &
  \begin{tabular}[c]{@{}l@{}}0.486\\0.496\\\textbf{0.026}*\end{tabular} &
  \begin{tabular}[c]{@{}l@{}}0.556\\0.569\\\textbf{0.028}*\end{tabular} \\ \hline
Wine &
  \begin{tabular}[c]{@{}c@{}}AP\\ IAPNA\\ APP\end{tabular} &
    \begin{tabular}[c]{@{}c@{}}0.492\\0.507\\\textbf{0.046}*\end{tabular} &
  \begin{tabular}[c]{@{}l@{}}0.563\\ 0.581\\\textbf{0.059}*\end{tabular} &
  \begin{tabular}[c]{@{}l@{}}0.639\\0.659\\\textbf{0.062}*\end{tabular} &
  \begin{tabular}[c]{@{}l@{}}0.719\\0.742\\\textbf{0.066}*\end{tabular} &
  \begin{tabular}[c]{@{}l@{}}0.804\\0.831\\\textbf{0.070}*\end{tabular} \\ \hline
Car &
  \begin{tabular}[c]{@{}c@{}}AP\\ IAPNA\\ APP\end{tabular} &
  \begin{tabular}[c]{@{}c@{}}1.215\\1.227\\\textbf{0.050}*\end{tabular} &
  \begin{tabular}[c]{@{}l@{}}1.325\\1.340\\\textbf{0.055}*\end{tabular} &
  \begin{tabular}[c]{@{}l@{}}1.440\\1.458\\\textbf{0.058}*\end{tabular} &
  \begin{tabular}[c]{@{}l@{}}1.559\\1.581\\\textbf{0.037}*\end{tabular} &
  \begin{tabular}[c]{@{}l@{}}1.684\\1.709\\\textbf{0.034}*\end{tabular} \\ \hline
KDD-CUP &
  \begin{tabular}[c]{@{}c@{}}AP\\ IAPNA\\ APP\end{tabular} &
  \begin{tabular}[c]{@{}c@{}}108.287\\108.928\\\textbf{2.207}*\end{tabular} &
  \begin{tabular}[c]{@{}l@{}}129.658\\130.381\\\textbf{2.850}*\end{tabular} &
  \begin{tabular}[c]{@{}l@{}}153.012\\153.819\\\textbf{3.029}*\end{tabular} &
  \begin{tabular}[c]{@{}l@{}}178.233\\179.128\\\textbf{3.207}*\end{tabular} &
  \begin{tabular}[c]{@{}l@{}}205.425\\ 206.408\\\textbf{3.400}*\end{tabular}
\end{tabular}%
}
\caption{Uniform-incremental experiment: comparison on Memory Usage. The highest score is denoted with an asterisk; the APP score is denoted in bold.}
\label{tab:comparison_mu}
\end{table}

\begin{table}[!ht]
\centering
\resizebox{0.90\columnwidth}{!}{%
\begin{tabular}{l|c|cllll}
\multicolumn{1}{c|}{Dataset} &
  Method & 1th & 2th & 3th & 4th & 5th \\ \hline
Iris &
  \begin{tabular}[c]{@{}c@{}}AP\\ IAPNA\\ APP\end{tabular} &
  \begin{tabular}[c]{@{}c@{}}59.0\\62.0\\\textbf{43.0}*\end{tabular} &
  \begin{tabular}[c]{@{}l@{}}49.0\\51.0\\\textbf{40.0}*\end{tabular} &
  \begin{tabular}[c]{@{}l@{}}164.0\\15.0*\\\textbf{50.0}\end{tabular} &
  \begin{tabular}[c]{@{}l@{}}156.0\\43.0*\\\textbf{43.0}*\end{tabular} &
  \begin{tabular}[c]{@{}l@{}}57.0\\37.0*\\\textbf{39.0}\end{tabular} \\ \hline
Wine &
  \begin{tabular}[c]{@{}c@{}}AP\\ IAPNA\\ APP\end{tabular} &
  \begin{tabular}[c]{@{}c@{}}60.0\\53.0\\\textbf{39.0}*\end{tabular} &
  \begin{tabular}[c]{@{}l@{}}55.0\\24.0*\\\textbf{40.0}\end{tabular} &
  \begin{tabular}[c]{@{}l@{}}63.0\\15.0*\\\textbf{41.0}\end{tabular} &
  \begin{tabular}[c]{@{}l@{}}61.0\\15.0*\\\textbf{39.0}\end{tabular} &
  \begin{tabular}[c]{@{}l@{}}65.0\\70.0\\\textbf{41.0}\end{tabular} \\ \hline
Car &
  \begin{tabular}[c]{@{}c@{}}AP\\ IAPNA\\ APP\end{tabular} &
  \begin{tabular}[c]{@{}c@{}}83.0\\15.0*\\\textbf{58.0}\end{tabular} &
  \begin{tabular}[c]{@{}l@{}}88.0\\127.0\\\textbf{43.0}*\end{tabular} &
  \begin{tabular}[c]{@{}l@{}}119.0\\34.0\\\textbf{15.0}*\end{tabular} &
  \begin{tabular}[c]{@{}l@{}}161.0\\166.0\\\textbf{41.0}*\end{tabular} &
  \begin{tabular}[c]{@{}l@{}}154.0\\15.0*\\\textbf{33.0}\end{tabular}\\ \hline
KDD-CUP &
  \begin{tabular}[c]{@{}c@{}}AP\\ IAPNA\\ APP\end{tabular} &
  \begin{tabular}[c]{@{}c@{}}103.0\\167.0\\\textbf{73.0}*\end{tabular} &
  \begin{tabular}[c]{@{}l@{}}115.0\\81.0\\\textbf{77.0}*\end{tabular} &
  \begin{tabular}[c]{@{}l@{}}133.0\\15.0*\\\textbf{70.0}\end{tabular} &
  \begin{tabular}[c]{@{}l@{}}142.0\\172.0\\\textbf{74.0}*\end{tabular} &
  \begin{tabular}[c]{@{}l@{}}139.0\\79.0\\\textbf{68.0}*\end{tabular} 
\end{tabular}%
}
\caption{Uniform-incremental experiment: comparison on the Number of Iterations. The highest score is denoted with an asterisk; the APP score is denoted in bold.}
\label{tab:comparison_ni}
\end{table}

\begin{table}[!ht]
\centering
\resizebox{0.90\columnwidth}{!}{%
\begin{tabular}{l|c|cllll}
\multicolumn{1}{c|}{Dataset} &
  Method & 1th & 2th & 3th & 4th & 5th \\ \hline
Iris$_{3}$ 
&
  \begin{tabular}[c]{@{}c@{}}AP\\ IAPNA\\ APP\end{tabular} &
  \begin{tabular}[c]{@{}c@{}}10.0\\5.0\\\textbf{4.0}*\end{tabular} &
  \begin{tabular}[c]{@{}l@{}}8.0\\6.0\\\textbf{3.0}*\end{tabular} &
  \begin{tabular}[c]{@{}l@{}}10.0\\5.0\\\textbf{3.0}*\end{tabular} &
  \begin{tabular}[c]{@{}l@{}}11.0\\7.0\\\textbf{3.0}*\end{tabular} &
  \begin{tabular}[c]{@{}l@{}}12.0\\9.0\\\textbf{2.0}*\end{tabular} \\ \hline
Wine$_{3}$ 
&
  \begin{tabular}[c]{@{}c@{}}AP\\ IAPNA\\ APP\end{tabular} &
  \begin{tabular}[c]{@{}c@{}}11.0\\9.0\\\textbf{4.0}*\end{tabular} &
  \begin{tabular}[c]{@{}l@{}}12.0\\1.0\\\textbf{2.0}*\end{tabular} &
  \begin{tabular}[c]{@{}l@{}}12.0\\2.0\\\textbf{3.0}*\end{tabular} &
  \begin{tabular}[c]{@{}l@{}}12.0\\2.0*\\\textbf{2.0}*\end{tabular} &
  \begin{tabular}[c]{@{}l@{}}12.0\\2.0\\\textbf{3.0}*\end{tabular} \\ \hline
Car$_{4}$
&
  \begin{tabular}[c]{@{}c@{}}AP\\ IAPNA\\ APP\end{tabular} &
  \begin{tabular}[c]{@{}c@{}}27.0\\25.0\\\textbf{8.0}*\end{tabular} &
  \begin{tabular}[c]{@{}l@{}}28.0\\26.0\\\textbf{4.0}*\end{tabular} &
  \begin{tabular}[c]{@{}l@{}}26.0\\25.0\\\textbf{2.0}*\end{tabular} &
  \begin{tabular}[c]{@{}l@{}}31.0\\29.0\\\textbf{50.0}*\end{tabular} &
  \begin{tabular}[c]{@{}l@{}}31.0\\28.0\\\textbf{3.0}*\end{tabular} \\ \hline
KDD-CUP$_{11}$
&
  \begin{tabular}[c]{@{}c@{}}AP\\ IAPNA\\ APP\end{tabular} &
  \begin{tabular}[c]{@{}c@{}}74.0\\4.0*\\\textbf{26.0}\end{tabular} &
  \begin{tabular}[c]{@{}l@{}}82.0\\6.0*\\\textbf{21.0}\end{tabular} &
  \begin{tabular}[c]{@{}l@{}}72.0\\6.0*\\\textbf{18.0}\end{tabular} &
  \begin{tabular}[c]{@{}l@{}}78.0\\63.0\\\textbf{16.0}*\end{tabular} &
  \begin{tabular}[c]{@{}l@{}}84.0\\72.0\\\textbf{20.0}*\end{tabular}
\end{tabular}%
}
\caption{Uniform-incremental experiment: comparison on the Number of Clusters. The highest score is denoted with an asterisk; the APP score is denoted in bold. The subscript denotes the number of categories in each dataset.}
\label{tab:comparison_nc}
\end{table}

The results show that APP achieves comparable/higher clustering performance than the conventional AP and IAPNA algorithms.
On average by considering all the time-steps and datasets, APP achieves a PUR score of 0.724, which is comparable but lower than 
the PUR score of AP (0.846) and IAPNA (0.755). This result can be explained by considering the number of clusters $NC$ created by the three algorithms, where we note that APP always returns the lowest value (see Table~\ref{tab:comparison_nc}). As a matter of fact, a high number of clusters positively affects the PUR metric without considering the possible noisiness of the created groups. On the opposite, APP achieves higher NMI score compared to AP and IAPNA. On average, APP obtains a NMI score of 0.553, while AP and IAPNA obtain 0.511 and 0.536, respectively. By considering the Wine and the Car datasets, we note that the NMI score of all the three algorithms is quite low. This is probably due to the 
categorical features in the such datasets that has been converted to numeric values by using one-hot encoding 
for vector representation. If we exclude the Wine and the Car dataset, the NMI average score of APP achieves the value of 0.726, while the AP and IAPNA scores are 0.647 and 0.657, respectively. As a further consideration, we note that the best results of APP in terms of NMI are reached on the KDD-CUP dataset where the average score is 0.731, while those of AP and IAPNA are 0.699 and 0.679, respectively. This is a particularly interesting result since KDD-CUP is the dataset with the highest number of objects and categories among those considered.

As a main result, due to the faithfulness property of APP that reduces the number of objects considered for clustering in each time-step, we observe that APP is far more scalable than AP and IAPNA in terms of CT, MU, and NI. On average by considering all the time-steps and datasets, APP achieves a CT score of 0.083, while AP and IAPNA achieve 8.633 and 14.017, respectively. Also about MU, we note that AP consumes 0.768 MB, while AP and IAPNA consume 39.359 MB and 39.573 MB, respectively. Furthermore, the average NI score of APP is 48.350, while AP and IAPNA obtain the score 101.300 and 62.800, respectively. According to the above results on the uniform-incremental experiment, we observe that APP is much faster than AP and IAPNA, while consuming much less memory than the two considered baselines. Furthermore, we note that the $NC$ values of APP represent the best approximation among the considered clustering algorithms with respect to the number of categories contained in the datasets. Usually, the $NC$ value of APP is slightly higher and sometimes equal to the number of dataset categories. \\

\subsubsection{Results on the variable-incremental experiment}
In the variable-incremental experiment, we performed the same tests of the uniform-incremental experiment on PUR, NMI, CT, MU, NI, and NC. For the sake of comparison, the scores of APP on all the tests and datasets of the variable-incremental experiment are shown in Table~\ref{tab:comparison_incremental}.  
\begin{table}[!ht]
\centering
\resizebox{0.9\columnwidth}{!}{%
\begin{tabular}{c|c|ccccc}
Dataset &
  Metric &
  1th &
  2th &
  3th &
  4th &
  5th \\ \hline
Iris$_3$ & 
  \begin{tabular}[c]{@{}c@{}}PUR\\ NMI\\ CT\\ MU\\ NI\\ NC\end{tabular} &
  \begin{tabular}[c]{@{}c@{}}1.000*\\ 0.616\\ 0.051\\ 0.016*\\ 59.0\\ 4.0\end{tabular} &
  \begin{tabular}[c]{@{}c@{}}0.988*\\ 0.696\\ 0.048\\ 0.020*\\ 45.0\\ 4.0\end{tabular} &
  \begin{tabular}[c]{@{}c@{}}0.938*\\ 0.751*\\ 0.051\\ 0.025\\ 51.0\\ 4.0\end{tabular} &
  \begin{tabular}[c]{@{}c@{}}0.897*\\ 0.754*\\ 0.048\\ 0.027\\ 46.0\\ 4.0\end{tabular} &
  \begin{tabular}[c]{@{}c@{}}0.887*\\ 0.718\\ 0.058\\ 0.038\\ 50.0\\ 5.0\end{tabular} \\ \hline
Wine$_3$
&
\begin{tabular}[c]{@{}c@{}}PUR\\ NMI\\ CT\\ MU\\ NI\\ NC\end{tabular} &
  \begin{tabular}[c]{@{}c@{}}0.816*\\ 0.412*\\ 0.058\\ 0.036*\\ 44.0*\\ 5.0\end{tabular} &
  \begin{tabular}[c]{@{}c@{}}0.823*\\ 0.518*\\ 0.044*\\ 0.048*\\ 39.5*\\ 4.0\end{tabular} &
  \begin{tabular}[c]{@{}c@{}}0.842*\\ 0.581*\\ 0.054\\ 0.057*\\ 39.5*\\ 4.0\end{tabular} &
  \begin{tabular}[c]{@{}c@{}}0.834*\\ 0.604*\\ 0.047*\\ 0.067\\ 37.0*\\ 3.0*\end{tabular} &
  \begin{tabular}[c]{@{}c@{}}0.742*\\ 0.572*\\ 0.047*\\ 0.079\\ 43.0\\ 5.0\end{tabular} \\
   \hline
Car$_4$
&
\begin{tabular}[c]{@{}c@{}}PUR\\ NMI\\ CT\\ MU\\ NI\\ NC\end{tabular} &
  \begin{tabular}[c]{@{}c@{}}0.770*\\ 0.364\\ 0.055*\\ 0.046*\\ 51.0*\\ 10.0\end{tabular} &
  \begin{tabular}[c]{@{}c@{}}0.677*\\ 0.323\\ 0.048*\\ 0.072\\ 43.0*\\ 11.0\end{tabular} &
  \begin{tabular}[c]{@{}c@{}}0.578\\ 0.278*\\ 0.037\\ 0.088\\ 46.0\\ 9.0\end{tabular} &
  \begin{tabular}[c]{@{}c@{}}0.604*\\ 0.315*\\ 0.034*\\ 0.084\\ 45.0\\ 10.0*\end{tabular} &
  \begin{tabular}[c]{@{}c@{}}0.535\\ 0.213\\ 0.032*\\ 0.100\\ 15.0*\\ 4.0\end{tabular} \\
   \hline
KDD-CUP$_{11}$
&
  \begin{tabular}[c]{@{}c@{}}PUR\\ NMI\\ CT\\ MU\\ NI\\ NC\end{tabular} &
  \begin{tabular}[c]{@{}c@{}}0.849*\\ 0.719\\ 1.804\\ 3.006\\ 87.5\\ 30.0\end{tabular} &
  \begin{tabular}[c]{@{}c@{}}0.838*\\ 0.732\\ 1.352\\ 3.629\\ 67.0*\\ 28.0\end{tabular} &
  \begin{tabular}[c]{@{}c@{}}0.831*\\ 0.737\\ 1.500\\ 4.054\\ 71.0\\ 28.0\end{tabular} &
  \begin{tabular}[c]{@{}c@{}}0.806*\\ 0.732*\\ 1.451\\ 4.584\\ 72.0*\\ 27.0\end{tabular} &
  \begin{tabular}[c]{@{}c@{}}0.744\\ 0.691\\ 1.479\\ 5.405\\ 64.0*\\ 25.0\end{tabular}
\end{tabular}%
}
\caption{Variable-incremental experiment: results of APP on all the considered datasets. The asterisks denote the APP scores higher than the corresponding ones in the uniform-incremental experiment.}
\label{tab:comparison_incremental}
\end{table}
As a general remark, we observe that the APP results on the variable-incremental experiment confirms the observations on the uniform-incremental experiment of Section~\ref{sec:evolutionary_scenario}. APP achieves comparable/higher clustering performances than AP and IAPNA algorithms. As a difference with the uniform-incremental experiment, we note that the PUR APP scores are improved. This is in relation with the fact that also a slightly higher number of clusters $NC$ are generated by APP in the variable-incremental experiment. The whole set of results on the variable-incremental experiment is available online at \url{https://github.com/umilISLab/APP}.\\

\paragraph{Ablation study} 
APP is designed to work under the ``group evolution'' assumption, namely the idea that a new incoming object that differs from past observations is more likely to be considered as an outlier of a previously created cluster rather than as a singleton new cluster. To this end, in the variable-incremental experiment, we inserted a $q$ parameter to specify the minimum number of incoming objects per category at a time-step $t$.
 
In the following, we present an ablation study, where the ``group evolution'' assumption is replaced by an ``individual evolution'' assumption. In particular, the constraint on the $q$ parameter is removed and it is possible that just one or few objects per category are incoming at a certain time-step $t$. The goal of this experiment is to analyse whether and how APP is capable of successfully recognising the category of incoming objects also when few elements of that category appear at a certain time-step.

In Table~\ref{tab:abl}, we show the APP results in terms of PUR and NMI when a minimum number of incoming objects per category $q$ is not specified/considered. 
\begin{table}[!th]
\centering
\resizebox{0.9\columnwidth}{!}{%
\begin{tabular}{c|c|ccccc}
Metric & Dataset & 1th & 2th & 3th & 4th & 5th \\ \hline
PUR    & \begin{tabular}[c]{@{}c@{}}Iris\\ Wine\\ Car\\ KDD-CUP99'\end{tabular} &  \begin{tabular}[c]{@{}c@{}}0.923 \\ 0.835* \\ 0.702 \\ \textbf{0.586}\end{tabular} & \begin{tabular}[c]{@{}c@{}} 0.900 \\  0.881* \\ 0.624 \\ \textbf{0.182} \end{tabular} &
\begin{tabular}[c]{@{}c@{}}   0.882 \\ 0.881* \\ 0.577 \\ \textbf{0.165}\end{tabular} &
\begin{tabular}[c]{@{}c@{}}  0.882\\ 0.889* \\ 0.602 \\ \textbf{0.135}\end{tabular} &\begin{tabular}[c]{@{}c@{}}  0.880 \\ 0.888* \\ 0.596* \\ \textbf{0.410}\end{tabular}  \\ \hline
NMI    & \begin{tabular}[c]{@{}c@{}}Iris\\ Wine\\ Car\\ KDD-CUP99'\end{tabular}  & \begin{tabular}[c]{@{}c@{}} 0.647*\\ 0.481*\\ 0.337 \\ \textbf{0.529} \end{tabular} &\begin{tabular}[c]{@{}c@{}} 0.677\\0.585* \\ 0.280 \\ \textbf{0.000} \end{tabular}& \begin{tabular}[c]{@{}c@{}}0.659\\ 0.629*\\ 0.240 \\ \textbf{0.000} \end{tabular}&\begin{tabular}[c]{@{}c@{}} 0.693\\ 0.642*\\ 0.288 \\\textbf{0.414} \end{tabular}& \begin{tabular}[c]{@{}c@{}}0.640\\0.615* \\ 0.300* \\\textbf{0.000}      \end{tabular} \\ \hline
\end{tabular}%
}
\caption{Ablation study: PUR and NMI scores of APP when the $q$ parameter is not considered and a minimum number of incoming objects per category is not employed. The APP scores that are higher with respect to Table~\ref{tab:comparison_incremental} are denoted with an asterisk; the scores on the KDD-CUP dataset are denoted in bold.}
\label{tab:abl}
\end{table}
With respect to the scores on PUR and NMI of Table~\ref{tab:comparison_incremental}, we note that the APP scores are slightly lower on Iris and Car datasets and they are slightly higher on the Wine dataset. We also note that the APP scores on the KDD-CUP dataset are dramatically lower than those shown in Table~\ref{tab:comparison_incremental}.

As a result, we argue that the ``group evolution'' assumption implemented through the $q$ parameter does not significantly affect the APP scores on small datasets like Iris, Car, and Wine where few categories are defined. On the opposite, on large datasets like KDD-CUP where a number of categories are defined, not using the $q$ parameter has a strong negative impact on PUR and NMI scores. This means that the ``group evolution'' assumption implemented through the $q$ parameter positively affects the correct recognition of object categories especially when datasets with several categories are considered, while not negatively affecting the PUR and NMI scores on datasets with few categories.\\

\begin{figure}[!ht]
\centering
\includegraphics[width=0.9\columnwidth]{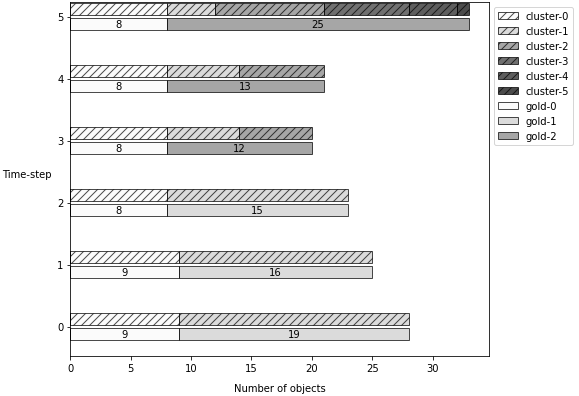}
\caption{Variable-incremental experiment: example of APP results by time-step over the Iris dataset.}
\label{fig:hist}
\end{figure}

\paragraph{Analysis of clustering results over time}
As a further test, we consider a specific execution of APP and the related clustering results over six time-steps. The goal is to analyse the capability of APP to correctly cluster objects according to the corresponding categories when different incremental schemas are used (i.e., growing, shrinking, stable). In Figure~\ref{fig:hist}, we show the results of an APP execution on the Iris dataset. 
In the dataset, the objects are distinguished in three different categories each one constituted by 50 elements, namely {\small\sf gold-0}, {\small\sf gold-1}, and {\small\sf gold-2}. In the test, the objects of the three categories follow a different incremental schema of arrival. The objects of the {\small\sf gold-0} category are scheduled for arrival according to the stable schema (i.e., 9 {\small\sf gold-0} objects at $0$-th and $1$-th time-steps; 8 {\small\sf gold-0} objects at subsequent time-steps). The objects of the {\small\sf gold-1} category follow a shrinking schema focused on time-steps from $0$-th to $2$-th. In particular, 19, 16, and 15 {\small\sf gold-1} objects are scheduled at $0$-th, $1$-th, and $2$-th time-steps, respectively. Finally the objects of the {\small\sf gold-2} category follow a growing schema focused on time-steps from $3$-th to $5$-th. In particular, 12, 13, and 25 {\small\sf gold-2} objects are incoming at $3$-th, $4$-th, and $5$-th time-steps, respectively.

In Figure~\ref{fig:hist}, for each time-step, we compare the clusters created by APP against the expected gold clusters based on the category of the incoming objects. We observe that APP works very well in clustering objects of stable and shrinking schemas. Indeed, the {\small\sf cluster-0} of APP always succeeds in correctly clustering the {\small\sf gold-0} objects in all the time-steps. Similarly, we note that the {\small\sf cluster-1} of APP perfectly reproduces the group of {\small\sf gold-1} objects in all the time-steps from $0$-th to $2$-th where the {\small\sf gold-1} objects are incoming. We also note that some incorrect clustering results are produced by APP on the {\small\sf gold-2} objects that arrive with a growing schema from $3$-th to $5$-th time-steps. In particular, in $3$-th and $4$-th time-steps, the {\small\sf gold-2} objects are distributed in two APP clusters, namely {\small\sf cluster-1} and {\small\sf cluster-2}. {\small\sf Cluster-2} represents the APP cluster that better fits to the {\small\sf gold-2} category. A part of the {\small\sf gold-2} objects are wrongly recognised as {\small\sf gold-1} objects and placed in {\small\sf cluster-1}. In the $5$-th time-step, the {\small\sf gold-2} objects are spread over five APP clusters. Again, a (small) part of {\small\sf gold-2} objects are placed in {\small\sf cluster-1} since they are wrongly recognised as {\small\sf gold-1} objects. Coherently with the results of $3$-th and $4$-th time-steps, the {\small\sf cluster-2} of APP seems to be the group that better fits the {\small\sf gold-2} category. The remaining {\small\sf cluster-3}, {\small\sf cluster-4}, and {\small\sf cluster-5} represent noisy groups with respect to the expected gold categories of Iris. According to the above observations, we argue that clustering errors mostly occur when the incoming objects follow a growing incremental schema. This is due to the fact that the new category appears with a low number of objects in the first time-step and this schema challenges the correct recognition of the new cluster to create.




\section{Application of APP to semantic shift detection}\label{sec:application}
As a concrete case-study of application of APP, we consider the semantic shift detection in the field of computational linguistics~\citep{montanelli2023survey,tahmasebi2018survey,kutuzov2018diachronic}. Semantic shift detection refers to the capability of recognising and measuring how much the use of a target word changes over time. Typically, given a target word $w$, the detection of a semantic shift in the use of $w$ is evaluated over two time-steps $t_1$ and $t_2$ characterised by distinct corpora of documents $C_1$ and $C_2$ where $w$ occurs. By generalizing such a scenario, semantic shift detection can be enforced over a sequence of time-steps $t_1, \dots, t_n$, each one associated with a bunch of documents and it can thus be analysed using APP, in that the documents (i.e., objects)  to consider are incrementally added and become available at different time-steps (i.e., dynamic arrival of objects). 
Furthermore, a number of occurrences with different meaning of the target word $w$ can appear in the documents arrived at a given time-step $t$ due to the possible polysemy of $w$. For instance, the word \verb|rock| is used with the meaning \verb|stone| in the sentences \verb|the tunnel was blasted out of solid rock| and \verb|they drilled through several layers of rock|. On the opposite, \verb|rock| is used with the meaning \verb|music| in the sentences \verb|John loves rock 'n roll| and \verb|He plays guitar in a rock band|. Clustering can be effectively employed over the documents of a certain time-step $t$ with the aim to create groups, each one containing the occurrences of the target word $w$ where a certain meaning of $w$ is employed. The comparison of groups calculated over different time-steps allows to recognise the possible shifts on the meanings of the word $w$.\\

\noindent\textbf{The Vatican case-study}.
As a case-study, we consider a diachronic corpus of Vatican publications and we focus on capturing how the meaning(s) of a target word changed over time~\citep{periti2022semantic}. The Vatican corpus contains 29k documents extracted from the digital archive of the Vatican website and it consists of all the web-available documents, spanning from the papacy of Eugene IV to Francis (1431-2023). Although the documents are available in various languages, including Italian, Latin, English, Spanish, and German, we downloaded the Italian corpus since a largest number of documents are available in this language. We note that the Vatican corpus is particularly appropriate for semantic shift detection since it is characterised by an exceptional historical depth and it deals with popular issues in the public debate, alongside themes of faith and worship.

To set-up the case-study, we first define a target word $w$ we aim to detect its semantic shift within the Vatican corpus. Then, we split the corpus in six sub-corpora, each one denoting a specific time period. It is worth noting that for most of the earlier pontificates, a few documents are available (e.g., Eugene IV) or none at all (e.g., Nicholas V). To address the skewed distribution of documents over time, we aggregated popes and related documents for ensuring that each sub-corpus contains at least 50 occurrences of the target word $w$. Furthermore, we performed a random sampling of 100 occurrences of $w$ from each sub-corpus when more occurrences are available to ensure that the number of occurrences are comparable across the sub-corpora.

To apply APP to the Vatican corpus, we follow the approach presented in~\citep{periti2022done}. In particular, we exploit the Italian pre-trained BERT model\footnote{dbmdz/bert-base-italian-cased} to represent each occurrence of the target word $w$ as a word embedding vector. The APP algorithm is then executed to create clusters of embeddings related to the same meaning of $w$. The first sub-corpus is considered in the initial run of APP, then the remaining sub-corpora are added one-by-one in a specific APP iteration. In the case-study, the pruning threshold $th_\gamma$ is set to $\infty$ since the goal of our experiment is to focus on the evolution of clusters over time, rather than to analyse the effects of the forgetfulness property on irrelevant clusters. 

\subsection{Cluster evolution analysis}
As a target word of our case-study, we consider $w=$ $\textsf{novit\`{a}}$ ($\textsf{novelty}$). The Vatican corpus is split into the following sub-corpora:  1) \textit{before Leo XIII}, with documents prior to 1878; 2) \textit{from Leo XIII to Pius XI}, with documents in the range 1878--1939; 3) \textit{from Pius XII to John XXIII}, with documents in the range 1939--1963; 4) \textit{Paul VI}, with documents in the range 1963--1978; 5) \textit{Benedict XVI}, with documents in the range 2005--2013; 6) \textit{Francis I}, with documents up to 2023. It is worth noting that we do not include the pontificate of John Paul II in this analysis. The richness and the variety of documents of John Paul II is significantly higher than the other pontificates and we note that it has been used in several different contexts and meanings, thus introducing a really challenging task of semantic shift detection. So, we decided to exclude the documents of John Paul II since the goal of our case-study is to show the behavior of APP on cluster evolution and not to discuss the APP effectiveness on a custom task of shift detection. As such, the effectiveness of APP for semantic shift detection will be discussed on a benchmark dataset in Section~\ref{sec:shiftsemeval}.

\begin{landscape}
\begin{figure}[!ht]
\centering
\includegraphics[width=\columnwidth]{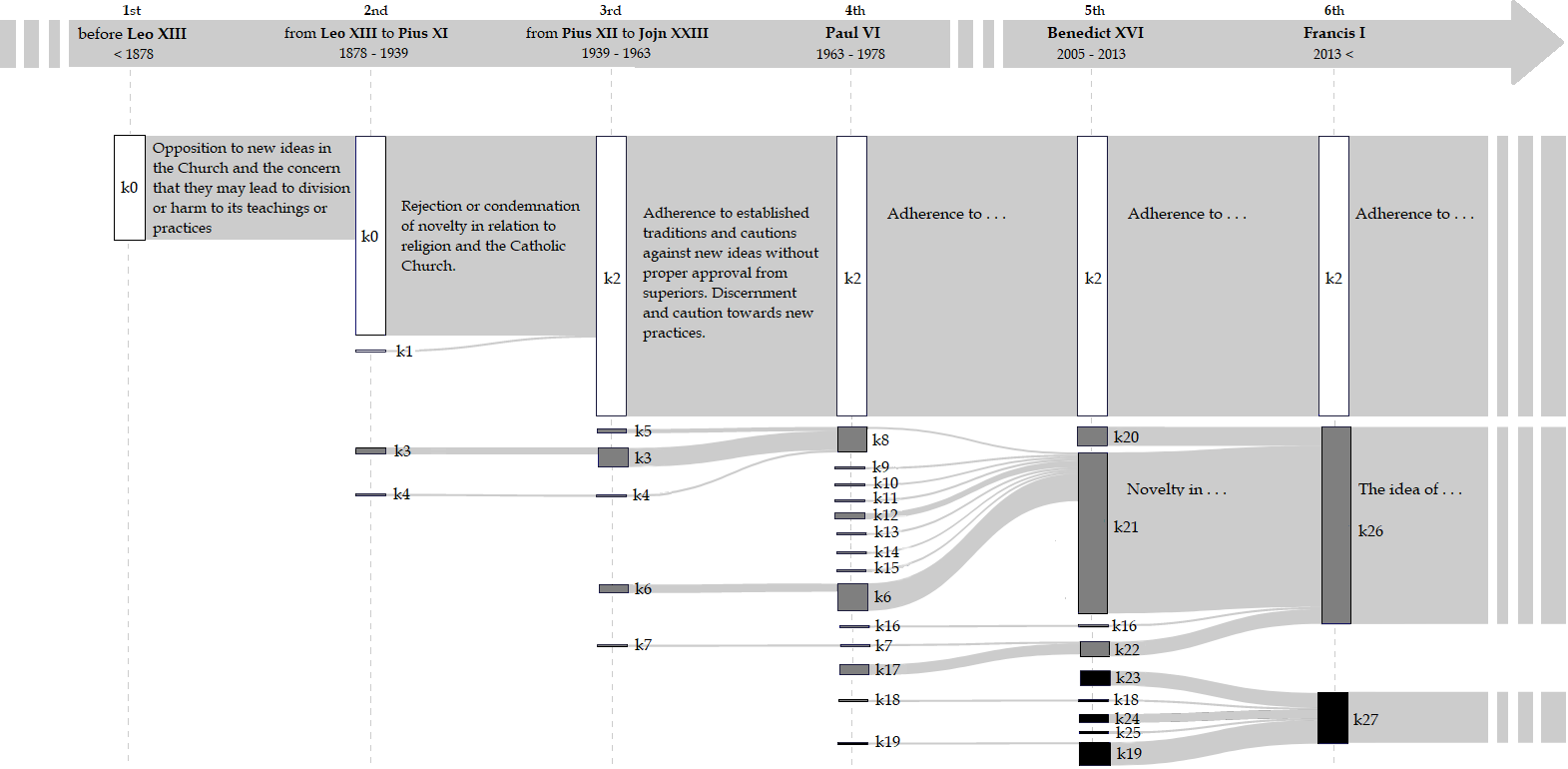}
\caption{The APP results on the Vatican corpus for the word $\textsf{novelty}$. 
}
\label{fig:stratification}
\end{figure}
\end{landscape}

In Figure~\ref{fig:stratification}, we provide an example of cluster evolution according to the stratification criteria presented in Section~\ref{sec:incremental}. 
Each cluster contains a set of contextual embeddings of the target word $\textsf{novelty}$ and it denotes a corresponding meaning of $\textsf{novelty}$ at a certain time by considering the documents of the Vatican corpus until that moment.

A cluster $\textsf{k}$ is represented as a box with an associated identifier. The cluster size denotes the cumulative number of elements in the cluster at each iteration: the larger the cluster box, the greater the number of cluster elements. In the example, we use the same cluster identifier across different iterations when the cluster is the result of a stratification-by-enrichment, while we assign new identifiers to clusters resulting from stratification-by-creation and stratification-by-merge.  

The example of Figure~\ref{fig:stratification} shows that just one meaning of the word $\textsf{novelty}$ could be recognised in the $1st$ APP iteration; and further meanings appeared in subsequent executions, especially in the iterations from $4th$ to $6th$, where the use of the word $\textsf{novelty}$ becomes strongly polysemous.

The cluster $\textsf{k0}$ in the $1st$ APP iteration is an example of stratification-by-creation and it describes the use of the word $\textsf{novelty}$ as a negative, dangerous concept, since new ideas and novel practices were considered as a threat to the traditional teachings of the Church by the earlier pontificates. The cluster $\textsf{k0}$ is populated with new elements in the $2nd$ iteration (stratification-by-enrichment), when a new cluster $\textsf{k1}$ is also introduced with embeddings of the $\textsf{novelty}$ occurrences from the documents of the $2nd$ sub-corpus (stratification-by-creation). The clusters $\textsf{k0}$ and $\textsf{k1}$ are joined in the $3rd$ iteration to generate the cluster $\textsf{k2}$ (stratification-by-merge). The cluster $\textsf{k2}$ remains unchanged in subsequent iterations from $4th$ to $6th$ (no more documents are found similar to $\textsf{k2}$), confirming that such a conservative, right-wing position of the Church has been abandoned after the Second Vatican Council (1962--1965). 

In this example, the clusters $\textsf{k0}$--$\textsf{k2}$ are equipped with a textual description that has the goal to summarise the cluster contents and the related meaning of the word $\textsf{novelty}$ in the cluster. Since cluster labeling is not the focus of this paper, we leverage ChatGPT\footnote{https://openai.com/blog/chatgpt/} to generate the cluster summaries of our examples. To label a cluster, we collect the text sources in the Vatican corpus that are associated with the occurrences of the word $\textsf{novelty}$ in the cluster and we ask ChatGPT to summarise the common topic. \\

\begin{landscape} 
\begin{figure}[!ht]
\centering
\includegraphics[width=0.8\columnwidth]{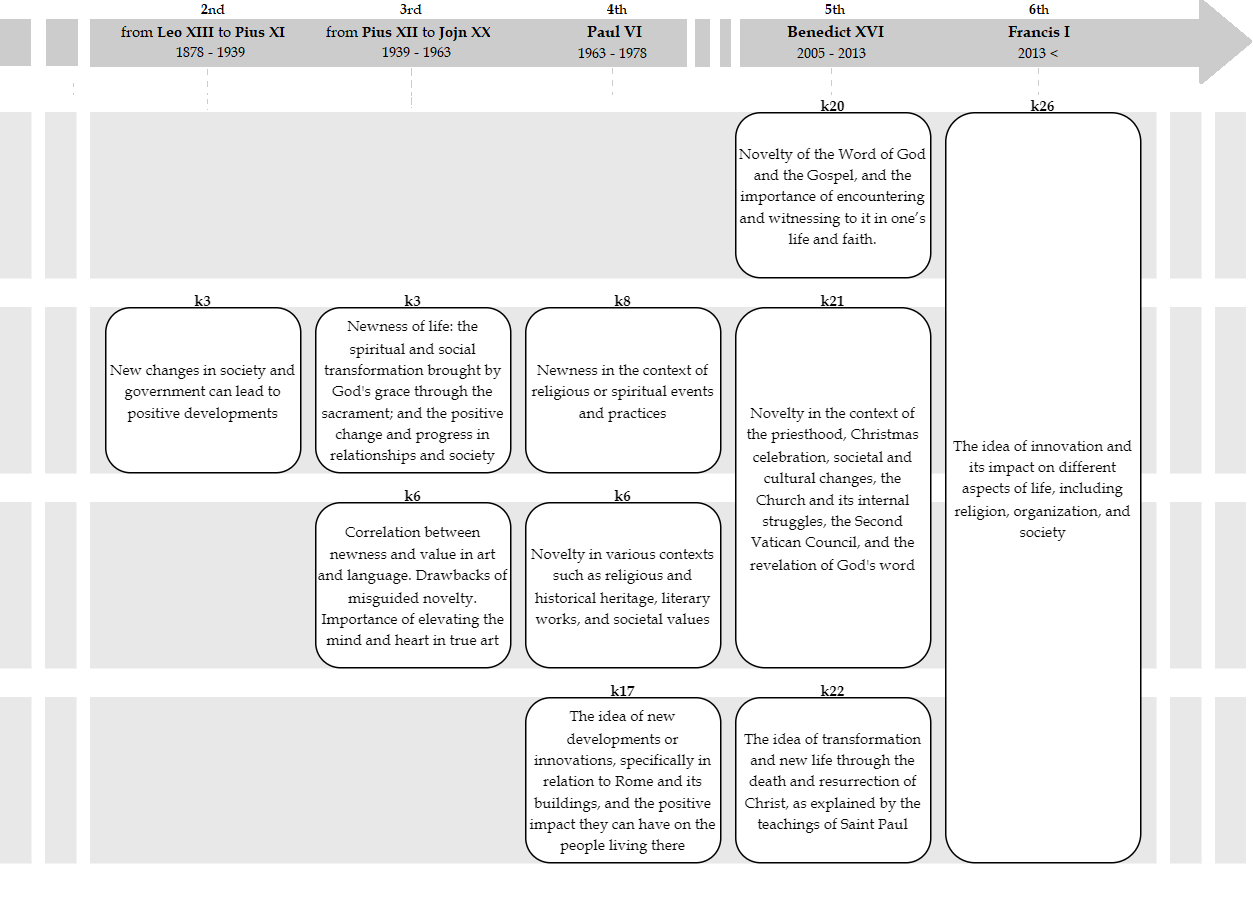}
\caption{The evolution/stratification of clusters that are finally merged into the cluster $\textsf{k26}$ of Figure~\ref{fig:stratification}. For the sake of readability, the cluster description is provided only for $\textsf{k3}$, $\textsf{k6}$, $\textsf{k8}$, $\textsf{k17}$, $\textsf{k20}$, $\textsf{k21}$, $\textsf{k22}$, $\textsf{k26}$.}
\label{fig:labeling}
\end{figure}
\end{landscape}

As a further example, in Figure~\ref{fig:labeling}, we show the evolution/stratification over time of those clusters that are finally merged into the cluster $\textsf{k26}$ at the $6th$ iteration of APP in Figure~\ref{fig:stratification}.
The example of Figure~\ref{fig:labeling} is about the usage of the word $\textsf{novelty}$ in relation with societal, cultural, and religious change. In particular, we focus on the period from 1939 to 2023 (iterations from $3rd$ to $6th$), although this meaning of $\textsf{novelty}$ appeared in the $2nd$ iteration with the clusters $\textsf{k3}$ and $\textsf{k4}$ as examples of stratification-by-creation. According to Figure~\ref{fig:stratification}, the $3rd$ iteration is characterised by the emergence of new relevant clusters such as $\textsf{k5}$ and $\textsf{k6}$ through stratification-by-creation, while the cluster $\textsf{k3}$ increases its importance with new elements through stratification-by-enrichment. The cluster $\textsf{k4}$ remains unchanged, and a new marginal cluster called $\textsf{k7}$ is created. In the $4th$ iteration, the number of clusters about this meaning of $\textsf{novelty}$ is strongly increased (stratification-by-creation), probably due to the dynamism of ideas introduced by the Second Vatican Council and reflected in the Vatican documents. Such a variety of positions at the $4th$ iteration is represented in Figure~\ref{fig:labeling} by the clusters $\textsf{k6}$, $\textsf{k8}$, and $\textsf{k17}$. The $5th$ iteration is mostly characterised by stratification-by-merge operations and the clusters $\textsf{k20}$, $\textsf{k21}$, and $\textsf{k22}$ represent the main result of APP on this meaning of $\textsf{novelty}$. About the cluster $\textsf{k21}$, we note that it is the result of a merge operation that involves a number of clusters of the previous iteration (i.e., the $4th$ one), and it is also strongly increased in importance due to the insertion of several elements (i.e., $\textsf{novelty}$ occurrences) of the current $5th$ iteration.

The result at the $5th$ iteration also includes the (minor) cluster $\textsf{k16}$ that remains unchanged with respect to the previous iteration (no elements of the $5th$ iteration are inserted in this cluster). The summary descriptions of clusters $\textsf{k20}$, $\textsf{k21}$, and $\textsf{k22}$ are provided in Figure~\ref{fig:labeling}. This meaning of $\textsf{novelty}$ is finally reconciled in a unique cluster $\textsf{k26}$ at the $6th$ iteration through a final stratification-by-merge operation.\\

\begin{figure*}[!ht]
\centering
\includegraphics[width=0.6\textwidth]{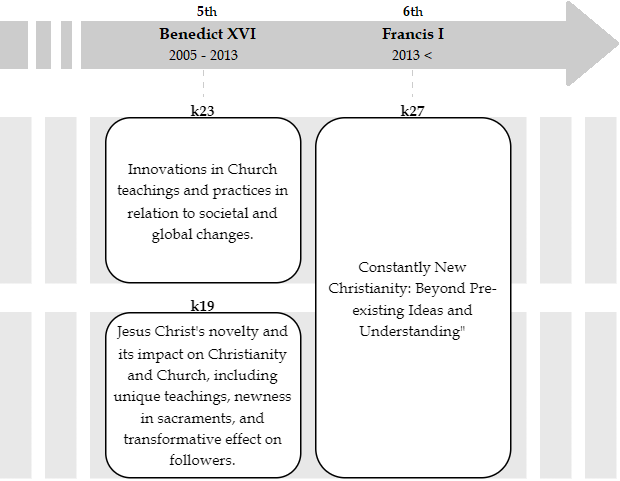}
\caption{The evolution/stratification of clusters $\textsf{k19}$, $\textsf{k23}$, and $\textsf{k26}$ of Figure~\ref{fig:stratification}.}
\label{fig:labeling1}
\end{figure*}

A final example of evolution/stratification is provided in Figure~\ref{fig:labeling1} about the clusters $\textsf{k19}$, $\textsf{k23}$, and $\textsf{k27}$ of Figure~\ref{fig:stratification}.
This example is about the usage of $\textsf{novelty}$ in relation with the innovation of the Christianity, new understanding of the Church teaching, and effects on the followers. In this example, we focus on the $5th$ and $6th$ iterations where most of the clusters about this meaning of $\textsf{novelty}$ appear, thus highlighting the very recent emergence of such a discussion in the Church debate. In Figure~\ref{fig:labeling1}, we show the descriptions of clusters $\textsf{k19}$ and $\textsf{k23}$ that are the most representative at the $5th$ iteration and that are finally merged into cluster $\textsf{k27}$ at the $6th$ iteration.\\

It is worth to stress that APP allows to represent all the various meaning/interpretations associated with the word $\textsf{novelty}$ at each iteration. Furthermore, the stratification criteria are able to track the transformations of clusters along the time, as well as to reconcile all the branches of a certain meaning into a summary cluster at the last iteration, thus providing a convenient picture to the scholar/analyst that aims to explore the evolution of $\textsf{novelty}$ in the whole Vatican corpus. 

\subsection{Evaluation on a reference benchmark}\label{sec:shiftsemeval}
As a final test for assessing the effectiveness of APP on semantic shift detection, we performed a benchmark evaluation by considering the Task 1 defined in the SemEval-2020 challenges~\citep{schlechtweg2020semeval}. The task is characterised by a number of corpora with related shifts to detect provided as gold standard. 

For application of APP to semantic shift detection, we extend the scheme proposed in~\citep{martinc2020capturing}. In particular, our approach relies on i) the use of contextualised embeddings to represent each occurrence of the target word from a BERT model~\citep{devlin2019bert}; ii) the aggregation of the embeddings with the APP clustering algorithm; iii) the computation of a semantic shift measure by comparing the vector distribution over clusters according to the time-steps by using the Jensen-Shannon divergence (JSD).\\

In Table~\ref{tab:comp_ss}, we report the best result we obtained on the SemEval Task 1 by considering the English and the Latin corpora. The results show the performance of AP, IAPNA, and APP in terms of Spearman's correlation. 
\begin{table}[!ht]
\centering
\begin{tabular}{l|ccc}
        & AP          & IAPNA          & APP \\ \hline
English & 0.514       & 0.462          & 0.512        \\
Latin   & 0.485       & 0.499          & 0.512        \\
\end{tabular}
\caption{Comparison of AP, IAPNA, and APP on SemEval Task 1 (further details are provided in~\cite{periti2022done}).}
\label{tab:comp_ss}
\end{table}
We observe that the results of APP are comparable and sometimes higher than AP and IAPNA. As occurred in both uniform-incremental and variable-incremental experiments, we also note that APP produces a smaller and more reasonable number of clusters compared to both AP and IAPNA. In particular, in the executed SemEval task, we found that the number of APP clusters generally varies between 0 and 30 while both AP and IAPNA produce more than 100 clusters, that is rather unrealistic if we consider that a cluster represents a word meaning. 

Further details about the APP results on the SemEval Task 1 for semantic shift detection are provided 
by~\cite{periti2023word,periti2022done}.

\section{Concluding remarks}\label{sec:conclusion}
In this paper, we propose A-Posteriori affinity Propagation (APP) as an extension of Affinity Propagation (AP). APP is conceived to work in incremental scenarios by enforcing faithfulness and forgetfulness through cluster consolidation/stratification. Evaluation results on popular benchmark datasets are provided to assess the performance of APP in two different incremental settings. The results show that APP obtains comparable results on cluster quality with respect to AP and IAPNA algorithms, while achieving high scalability performances at the same time. Further experimental results about the use of APP for semantic shift detection are discussed to highlight the applicability of our algorithm to a concrete evolutionary scenario. More in general, APP is suitable for application scenarios where the ``group evolution'' assumption holds, like for example tracking the evolution of word meanings over diachronic corpora. Further application scenarios of APP are in the field of Computational Linguistics and Natural Language Processing where the use of multi-dimensional vector representations (e.g., the 768-dimensional BERT embeddings) is getting popular for representing the semantics of words and sentences. In this case, the average-based representation of cluster centroids enforced by APP is particularly appropriate to manage the embeddings generated by the modern large deep language models in a scalable way.

\printcredits

\bibliographystyle{cas-model2-names}

\bibliography{cas-refs}

\bio{}
\endbio

\end{document}